\documentclass[11pt]{article}

\PassOptionsToPackage{hyperfootnotes=false}{hyperref}
\usepackage[final]{acl}
\usepackage{tocloft}
\usepackage{times}
\usepackage{latexsym}
\usepackage[T1]{fontenc}
\usepackage[utf8]{inputenc}
\usepackage{microtype}
\usepackage{inconsolata}

\usepackage{graphicx}
\usepackage{wrapfig}
\usepackage{multirow}
\usepackage{booktabs}
\usepackage{amsmath,amssymb}
\usepackage{tabularx}
\usepackage{makecell}
\usepackage{pifont}
\usepackage[most]{tcolorbox}
\tcbuselibrary{breakable}

\definecolor{mybluebg}{rgb}{0.9, 0.95, 1.0}
\definecolor{myblueframe}{rgb}{0.1, 0.2, 0.6}
\definecolor{softblue}{RGB}{100, 149, 237}
\definecolor{darkpurple}{RGB}{160, 0, 160}
\definecolor{darkred}{RGB}{160, 0, 0}
\definecolor{darkblue}{RGB}{0, 0, 160}
\definecolor{OliveGreen}{RGB}{34,139,34}
\definecolor{BrickRed}{RGB}{203, 65, 84}

\newtcolorbox{infobox}[2][]{breakable,#1,
    enhanced,
    colback=mybluebg,
    colframe=myblueframe,
    coltitle=black,
    fonttitle=\bfseries\large,
    halign title=center,
    boxrule=0.5pt,
    arc=3mm,
    attach boxed title to top center={yshift=-2mm},
    boxed title style={
        colback=mybluebg,
        boxrule=0pt,
        arc=0mm,
    },
    underlay={
        \draw[myblueframe, line width=0.5pt]
            (title.south west) -- (title.south east);
    },
    title=#2,
    #1
}

\tcbset{
  colframe=gray!70!black,
  colback=gray!5!white,
  boxrule=0.3pt,
  width=\textwidth,
  arc=1mm,
  fonttitle=\bfseries,
  coltitle=white,
}

\newcommand{\cmark}{\color{OliveGreen}\ding{51}}%
\newcommand{\xmark}{\color{BrickRed}\ding{55}}%

\title{LLM as GNN: Graph Vocabulary Learning for Text-Attributed Graph Foundation Models}

\author{
  \textbf{Xi Zhu}\textsuperscript{1}\thanks{Equal contribution. Email: \texttt{xi.zhu@rutgers.edu} } \quad
  \textbf{Haochen Xue}\textsuperscript{2}\footnotemark[1] \quad
  \textbf{Ziwei Zhao}\textsuperscript{3} \quad
  \textbf{Wujiang Xu}\textsuperscript{1} \quad
  \textbf{Jingyuan Huang}\textsuperscript{1}
  \\
  \textbf{Minghao Guo}\textsuperscript{1} \quad
  \textbf{Qifan Wang}\textsuperscript{4} \quad
  \textbf{Kaixiong Zhou}\textsuperscript{5} \quad
  \textbf{Imran Razzak}\textsuperscript{2} \quad
  \textbf{Yongfeng Zhang}\textsuperscript{1}
  \\
  \textsuperscript{1}Rutgers University \quad
  \textsuperscript{2}Mohamed bin Zayed University of Artificial Intelligence (MBZUAI)
  \\
  \textsuperscript{3}University of Science and Technology of China \quad
  \textsuperscript{4}Meta AI \quad
  \textsuperscript{5}North Carolina State University
}

\begin{document}
\maketitle
\begin{abstract}
 Text-Attributed Graphs (TAGs), where each node is associated with text descriptions, are ubiquitous in real-world scenarios. They typically exhibit distinctive structure and domain-specific knowledge, motivating the development of Graph Foundation Models (GFMs) that generalize across diverse graphs and tasks. Despite large efforts to integrate Large Language Models (LLMs) and Graph Neural Networks (GNNs) for TAGs, most approaches suffer from decoupled architectures with two-stage alignment, limiting their synergistic potential. Even worse, existing methods assign out-of-vocabulary (OOV) tokens to graph nodes, leading to graph-specific semantics, token explosion, and incompatibility with task-oriented prompt templates, which hinders cross-graph and cross-task transferability. To address these challenges, we propose PromptGFM, a versatile GFM for TAGs grounded in graph vocabulary learning. PromptGFM comprises two key components: (a) a graph understanding module, which explicitly prompts LLMs to replicate the fine-grained GNN workflow within the text space, facilitating seamless GNN-LLM integration and elegant graph-text alignment; and (b) a graph inference module, which establishes a language-based graph vocabulary to ensure expressiveness, transferability, and scalability, enabling readable instructions for joint LLM fine-tuning across graphs and tasks. Extensive experiments demonstrate the superiority of PromptGFM, highlighting its strong transferability across diverse graphs and tasks and its adaptability to auxiliary pre-training tasks.
 The code is available at \url{https://github.com/agiresearch/PromptGFM}.
\end{abstract}

\section{Introduction}
\label{sec: intro}

Graphs, characterized by their non-Euclidean structures and rich domain-specific knowledge, serve as fundamental representations of complex relational data. Many real-world graphs are further enriched with node-level textual attributes, forming Text-Attributed Graphs (TAGs), such as citation networks ~\citep{DBLP:journals/ipm/Eto19, DBLP:journals/qss/BunemanDLS21}, social networks ~\citep{DBLP:conf/kdd/KempeKT03,DBLP:conf/www/MyersSGL14}, and molecular graphs ~\citep{wieder2020compact,DBLP:journals/corr/abs-2405-06649}. However, existing methods often depend on task- or dataset-specific training, limiting their transferability. This motivates the development of Graph Foundation Models (GFMs) that generalize across graphs and tasks~\citep{DBLP:conf/icml/MaoCT000S0T24,DBLP:conf/emnlp/XiaK024,sun2025graphicl}.

\begin{figure}[!tbp]
    \centering
    \includegraphics[width=0.5\textwidth]{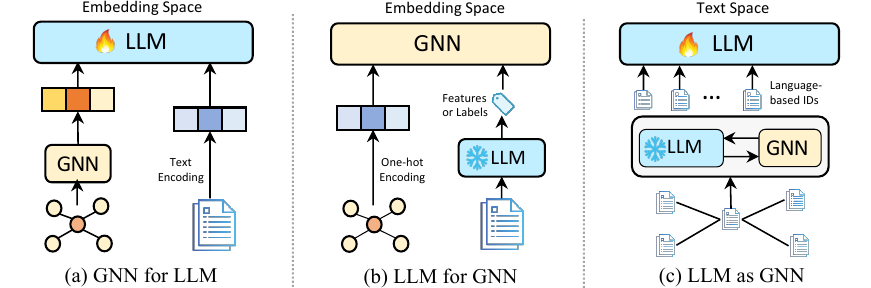}
    \caption{Overview of three GNN-LLM integration paradigms for graph-text alignment: (a) \textit{GNN for LLM} and (b) \textit{LLM for GNN} use decoupled architectures in the embedding space, while (c) our work operates \textit{LLM as GNN} in the text space. }
    \label{Motivation_1}
\end{figure}

In TAGs, GNNs and LLMs are commonly combined to process structural and textual modalities, as illustrated in Figure~\ref{Motivation_1}. 
\textbf{(a) GNN for LLM.} GNNs produce structure-aware node embeddings that complement textual embeddings and improve LLM inference~\citep{DBLP:conf/sigir/Tang00SSCY024, DBLP:journals/corr/abs-2310-05845, DBLP:conf/www/0011H0C24}. 
\textbf{(b) LLM for GNN.} LLMs extract semantic features or generate pseudo-labels from text to supervise GNN training~\citep{DBLP:conf/iclr/ChenMWH0Z0T24, DBLP:conf/iclr/0057FKLT0Z24, DBLP:conf/ijcai/ZhuWST24}. 
However, these two-stage pipelines only loosely couple GNNs and LLMs, and thus struggle to fully exploit their complementary strengths, often resulting in suboptimal graph-text alignment.

\begin{figure}[!tbp] 
    \centering
    \caption{LLM-driven replication of the GNN workflow. We achieve fine-grained alignment between traditional embedding-based GNN and our prompt-based GNN.}
    \label{Motivation_2}
    \includegraphics[width=0.45\textwidth]{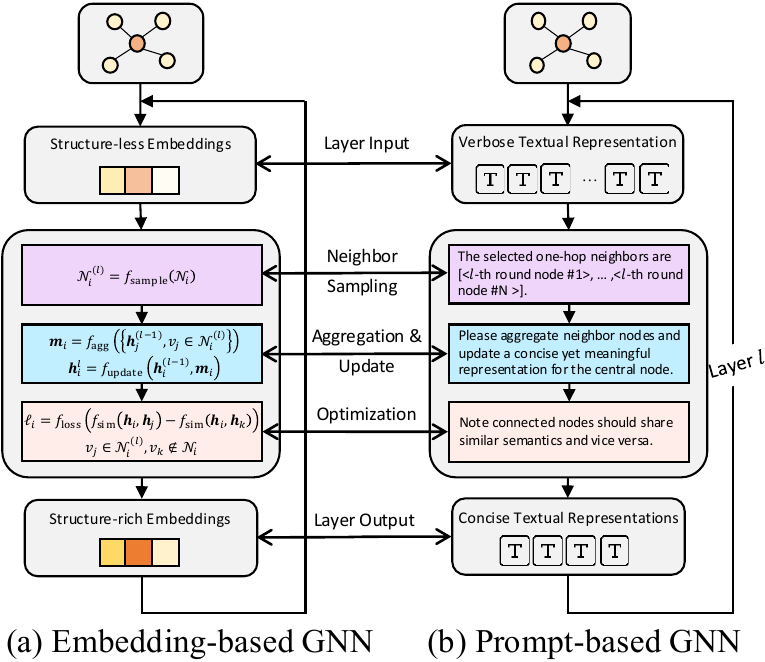}
\end{figure}

Recently, a growing trend explores \textbf{LLM as GNN}, where graph verbalizers convert graph data into heuristic or code-like prompts, enabling LLMs to capture graph semantics and structure~\citep{DBLP:conf/eacl/YeZWXZ24, DBLP:conf/acl/WangWH00M24, DBLP:conf/icml/Chen0JSW24}. 
However, we argue that these methods are not true GNNs, as they omit the core message-passing mechanism. 
As shown in Figure~\ref{Motivation_2}(a), a standard GNN layer typically involves neighbor sampling, aggregation-update, and optimization~\citep{DBLP:conf/iclr/KipfW17,DBLP:journals/corr/abs-1710-10903}. 
Stacking such layers transforms structure-less embeddings into structure-rich representations with high-order dependencies. 
This raises a key question: \textbf{\textit{Can LLMs replicate the GNN workflow in text space to jointly capture graph semantics and structure?}}

Meanwhile, existing works intuitively treat each graph node as an out-of-vocabulary (OOV) token, resulting in graph-specific semantics and thus uncontrolled vocabulary explosion~\citep{DBLP:conf/sigir/Tang00SSCY024, DBLP:conf/acl/ZhouDZ0XXH25}. Worse still, due to vocabulary mismatches, ID-based node embeddings and language-based token embeddings reside in different feature spaces, causing semantic misalignment during LLM inference. This incompatibility undermines both transferability and scalability of graph-specific knowledge across diverse graphs and tasks. To enable knowledge transfer, another key question emerges: \textbf{\textit{Can OOV tokens be replaced with compatible and universal node representations to build a versatile and transferable GFM?}}

Inspired by prior work~\citep{DBLP:conf/icml/Cai24, DBLP:conf/icml/MaoCT000S0T24, DBLP:conf/nips/WangZCZ024}, we argue that a true GFM should be grounded in \textbf{{graph vocabulary learning}} with the following properties: \textbf{{(1) Expressiveness.}} It encapsulates both semantic and structural information across diverse graphs. \textbf{(2) Transferability.} Every node in any graph should be representable by a composition of fundamental units within the vocabulary. \textbf{(3) Scalability.} The vocabulary should be inclusive to accommodate unseen nodes, even beyond existing graphs. 
Since natural language inherently provides expressive and transferable tokens~\citep{DBLP:journals/jmlr/RaffelSRLNMZLL20, DBLP:conf/icml/RadfordKHRGASAM21, DBLP:journals/corr/abs-2307-09668, DBLP:conf/nips/WangZCZ024}, we establish a universal graph vocabulary in the text space, and then propose \textbf{PromptGFM}, a GFM for TAGs with the following components.




\textbf{Graph Understanding Module.} 
To operate LLMs as GNNs, we initialize node features with textual attributes and prompt LLMs to replicate the fine-grained GNN workflow in text space. As Figure \ref{Motivation_2} shows, we encode local structure via one-hop neighbor descriptions and design prompts for a flexible aggregation-update mechanism. To support optimization, we introduce heuristic prompts inspired by contrastive learning at each layer. Iterative LLM calls then simulate message passing over the graph, progressively refining verbose textual features into concise yet informative textual representations. Unlike traditional GNNs that output numerical embeddings, our prompt-based GNN produces structured textual encodings that preserve node semantics and higher-order dependencies. Consequently, LLMs can operate as GNNs, and GNNs can be interpreted as LLMs, unlocking the full potential of GNN-LLM integration and empowering elegant graph-text alignment.

\textbf{Graph Inference Module.} 
After capturing semantics and structure with prompt-based GNNs, we decouple the resulting textual representations to establish a graph vocabulary, mapping each node to a finite sequence of language-based tokens, essentially as language-based IDs. 
Because these IDs are drawn from the LLM's native vocabulary, they preserve semantic fidelity and avoid OOV issues. 
Thus, this vocabulary is universally interpretable, transferable across graphs, and scalable to unseen nodes, resolving semantic irrelevance and graph-text incompatibility.
These language-based IDs are then integrated into readable natural-language prompts. 
To support generalization, we adopt multi-instruction fine-tuning over diverse graphs and tasks, enabling cross-graph and cross-task knowledge transfer. 
Overall, this graph vocabulary not only mitigates graph-text incompatibility but also provides a foundation for general GFMs.
Our contributions are summarized as follows:

\noindent $\bullet$ We present PromptGFM, a graph foundation model for TAGs built on graph vocabulary learning.

\noindent $\bullet$ We introduce an LLM-as-GNN paradigm in text space, enabling seamless GNN-LLM integration and effective graph--text alignment.

\noindent $\bullet$ We construct a language-based graph vocabulary as the cornerstone of the GFM, ensuring graph-text semantic fidelity and eliminating the OOV issue.

\noindent $\bullet$ We conduct extensive experiments to validate the superiority of PromptGFM, especially its cross-graph and cross-task transferability and adaptability to auxiliary pre-training tasks.

\section{Preliminaries}

\textbf{Text-Attributed Graphs (TAGs).} 
A TAG is defined as ${G}=\left(V,E, X\right)$, where $V$ is the node set, $E$ is the edge set, and $X$ is the set of textual node attributes. Each node $v_i \in {V}$ is associated with a textual description ${X}_i=\left(x_{i}^{1}, x_{i}^{2}, \ldots, x_{i}^{{n_i}} \right)$, where each $x_{i}^{k} \in \mathcal{X}, k = 1, \dots, {n_i}$, and $\mathcal{X}$ denotes the natural language dictionary.

\noindent \textbf{Graph Neural Networks.} GNNs are state-of-the-art models for graph machine learning, built on the message-passing mechanism. In practice, a GNN selects the neighboring nodes of the target node, aggregates their representations to capture local structure, and updates the target node's representation. Mathematically, for a central node $v_i$, the $l$-th layer of a general GNN is defined as:
\begin{equation}
\begin{aligned}
    \mathcal{N}_i^{(l)} &= f_{\text{sample}}\left(\mathcal{N}_i\right), \\
    \mathbf{m}_i^{(l)} &= f_{\text{agg}}\left({\left \{\mathbf{h}_j^{(l-1)}, v_j \in \mathcal{N}_i^{(l)}\right \} }\right),\\
    \mathbf{h}_i^{(l)} &= f_{\text{update}}\left(\mathbf{h}_i^{(l-1)}, \mathbf{m}_i^{(l)}\right)\\
\end{aligned}
\end{equation}
\noindent where $\mathbf{h}_i^{(l)}$ is the node embedding of $v_i$ at the $l$-th layer. $\mathcal{N}_i$ denotes the full neighbor set and $\mathcal{N}_i^{(l)}$ the sampled neighbor set at the $l$-th layer. To capture higher-order dependencies, we stack $L$ layers and derive final embeddings as $\mathbf{h}_i = f_{\text{pooling}}\left(\mathbf{h}_i^{(1)}, ..., \mathbf{h}_i^{(L)}\right)$ ~\citep{DBLP:journals/tnn/GrattarolaZBA24}. For optimization, a contrastive loss with negative sampling is commonly used in unsupervised graph learning ~\citep{DBLP:conf/nips/HamiltonYL17, DBLP:conf/iclr/VelickovicFHLBH19}, expressed as:
\begin{equation}
    \begin{split}
        \ell = f_{\text{loss}}\left(
        f_{\text{sim}}\left(\mathbf{h}_i, \mathbf{h}_j\right), f_{\text{sim}}\left(\mathbf{h}_i, \mathbf{h}_k\right)\right),
    \end{split}
\end{equation}
\noindent where $v_j \in \mathcal{N}_i^{(l)}$ is a positive sample and $v_k \notin \mathcal{N}_i$ is a negative one. $f_{\text{sim}}\left(\cdot \right)$ measures the similarity between two nodes. $f_{\text{loss}}\left(\cdot \right)$ enforces contrastive learning by increasing the similarity for connected nodes and reducing it for unconnected ones.

\section{Methodology}

\begin{figure*}[!tbp]
    \centering
    \vspace{-2mm}
    \caption{The pipeline of PromptGFM. (a) Graph Understanding Module: For arbitrary TAGs from different domains, prompt-based GNN replicates traditional embedding-based GNN workflow in the text space, generating compact node representations. (b) Graph Inference Module: We establish a unified graph vocabulary and extract language-based IDs to generate massive pure-language prompts, enabling LLM fine-tuning across graphs and tasks to support knowledge transfer and flexible adaptability.}
    \label{Fig_Pipeline}
    \includegraphics[width=0.93\textwidth]{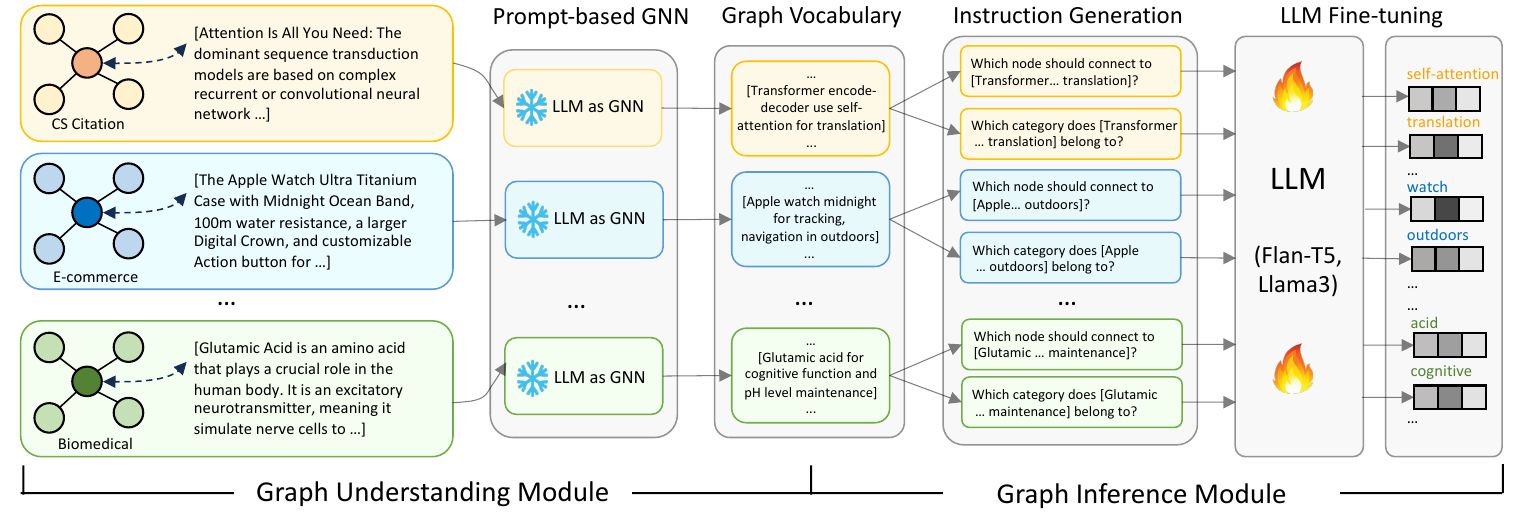}
    \vspace{-3mm}
\end{figure*}

We illustrate the pipeline of our proposed PromptGFM in Figure ~\ref{Fig_Pipeline}, which consists of a graph understanding module and a graph inference module.


\subsection{Graph Understanding Module}

The graph understanding module generates expressive and generalizable node representations for downstream inference. 
The key challenge lies in aligning semantic and structural information, as LLMs excel at textual understanding while GNNs specialize in structural modeling~\citep{DBLP:journals/corr/abs-2311-12399, DBLP:conf/kdd/RenTYCH24}. 
To bridge this gap, we propose prompt-based GNNs that prompt LLMs to replicate the general GNN workflow in text space, which requires appropriate prompts guided by three factors: 
\textbf{(a) Graph Structure.} How to simulate message passing to encode structural dependencies? 
\textbf{(b) Node Semantics.} How to distill essential semantics into concise yet informative representations?  
\textbf{(c) Graph Description.} How to convey graph semantics and structure to LLMs simultaneously?

As illustrated in Figure \ref{Motivation_2}, we implement a fine-grained replication of a general GNN workflow, i.e., the \textbf{prompt-based GNN}. We first initialize node representations by summarizing textual attributes, analogous to the embedding lookup in conventional GNNs. To reduce computational overhead, we sample one-hop neighbors as follows:
\begin{equation} 
\label{Neighbor Sampling.}
\{X_j^{(l-1)}, v_j \in \mathcal{N}_i^{(l)}\}
\leftarrow \text{Prompt}_{\text{sample}}(X^{(l-1)}, \mathcal{N}_i)
\end{equation}

\noindent where $X^{(l-1)} = \left \{X_0^{(l-1)}, X_1^{(l-1)}, \dots, X_{|V|-1}^{(l-1)} \right \}$ are the textual representations of all nodes from the previous layer, and $\mathcal{N}_i^{(l)} \subseteq \mathcal{N}_i$ denotes the neighbors sampled for $v_i$ at the $l$-th layer. Then, we directly prompt LLMs to perform message aggregation and update for each node as follows:
\begin{equation}
\label{Agg-Update.}
\begin{aligned}
X_i^{(l)} \leftarrow \text{Prompt}_{\text{agg-update}}
(&\{X_j^{(l-1)}, v_j \in \mathcal{N}_i^{(l)}\},\\
&X_i^{(l-1)}),
\end{aligned}
\end{equation}
\noindent where we seek a flexible message-passing mechanism without relying on predefined operators such as mean and weighted aggregators. We analyze its sensitivity to input order in Appendix~\ref{sec: permutation} and discuss prompt length constraints in Appendix~\ref{Prompt Length Limitation Discussion}. 
In terms of optimization, we follow unsupervised graph learning and adopt a prompt-driven contrastive objective, which encourages similarity among neighboring nodes while pushing apart distant ones, serving as a soft structural prior that guides the LLM to filter and integrate informative neighbor signals. As a task-agnostic approach to graph representation learning, contrastive optimization naturally aligns with the generality required by GFMs.
While traditional negative sampling becomes redundant in our setting, we express this objective as natural-language instructions in the prompt, without introducing an explicit training loss. Prompt details are provided in Appendix~\ref{Prompt Design}.

After repeating $L$ rounds, we obtain the final textual representation as $T_i = X_i^{(L)}$, which encodes both semantic and structural information and addresses the aforementioned challenges: \textbf{(a) Graph Structure.} We iteratively invoke LLMs with shared prompt templates, feeding each round's output into the next to capture higher-order dependencies, and trace this process for individual nodes in Appendix~\ref{Case Study}. \textbf{(b) Node Semantics.} Simultaneously, we explicitly instruct LLMs to produce concise yet expressive node representations, gradually refining them into denser and richer semantic descriptions. \textbf{(c) Graph Description.} One-hop neighbor descriptions collectively encode the adjacency structure, representing the entire graph while remaining efficient (\S\ref{Computational Complexity}). We further investigate layer-wise representation discrimination, semantic retention, and module performance in Appendix~\ref{Prompt-based GNN Analysis}. This workflow-level replication in text space exemplifies the potential of treating LLMs as GNNs for GNN-LLM integration and graph-text alignment.

\begin{figure*}[!tbp]
    \vspace{-2mm}
    \centering
    \caption{An instance of graph inference module in link prediction, where language-based IDs are indexed from the graph vocabulary to generate readable instructions with task-oriented templates. A multi-instruction fine-tuning framework is employed to unify graphs and tasks and learn transferable knowledge for GFMs.}
    \includegraphics[width=0.95\textwidth]{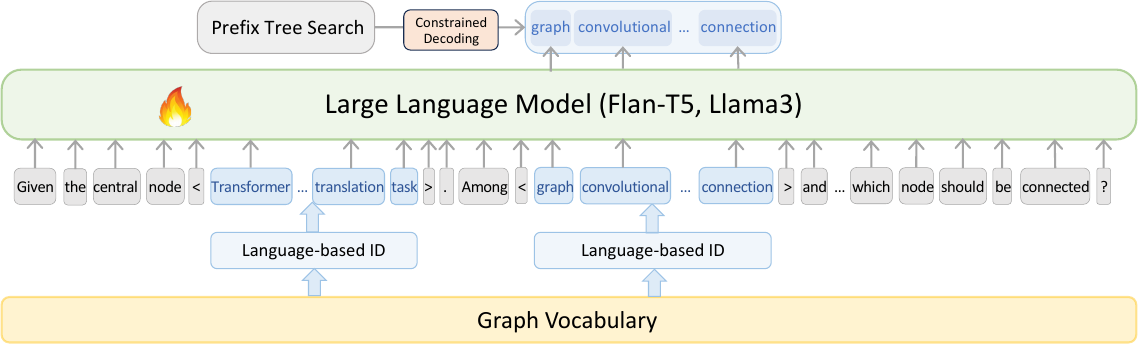}
    \label{Fig_GIM}
    \vspace{-2mm}
\end{figure*}

\subsection{Graph Inference Module}

The graph inference module seeks to unify diverse graphs and tasks and acquire transferable knowledge through multi-instruction LLM fine-tuning. Existing methods, which represent nodes as OOV tokens, create semantic misalignment between ID-based and language-based embeddings in task-oriented prompts due to distinct vocabularies, and substantially limit dataset-level transferability and task-level adaptability ~\citep{DBLP:conf/sigir/Tang00SSCY024, DBLP:conf/iclr/XiaZHG0LLL23, DBLP:conf/eacl/YeZWXZ24}. To overcome this challenge, we introduce a language-based graph vocabulary that bridges this incompatibility, enabling the generation of large-scale, coherent, and readable instructions to support effective LLM inference.

\noindent \textbf{Graph Vocabulary.}
A transferable graph vocabulary, whose fundamental units can represent each node, is essential for effective GFMs and should satisfy three key criteria: \textbf{expressiveness}, \textbf{transferability}, and \textbf{scalability}. 
Since each node has been equipped with a textual representation encoding its core semantics and local structure, we aggregate these representations to build a universal language-based graph vocabulary compatible with the LLM's native vocabulary. In this vocabulary, each node is then represented by a finite sequence of language tokens, called a \textbf{language-based ID}, defined as:
\begin{equation}
\begin{aligned}
    \mathcal{F}: V \rightarrow T,
\end{aligned}
\end{equation}
\noindent where each node $v_i$ is mapped to a sequence ${T}_i=\left(t_{i}^{1}, t_{i}^{2}, \ldots, t_{i}^{m_i} \right)$, with $t_i^k \in \mathcal{X}$ and $\mathcal{X}$ denoting the dictionary of natural language tokens. In practice, $T_i$ is simply the tokenized form of the final textual representation produced by the graph understanding module, requiring no additional summarization, clustering, or remapping. Collectively, the language-based IDs of all nodes in $V$ are denoted as $T = \left\{T_0, T_1, \dots, T_{|V|-1}\right\}$.
We analyze vocabulary coverage and uniqueness in Appendix~\ref{sec: vocabulary_analysis}. 
Along this line, the structured token sequence of a graph node functions like a word in a standard lexicon, preserving semantic fidelity and eliminating the OOV issue.
Overall, our graph vocabulary satisfies the desired criteria:
\textbf{(a) Expressiveness.} Language-based IDs preserve rich semantic and structural information from open-world settings.
\textbf{(b) Transferability.} Like human lexicons, the vocabulary is grounded in natural language, providing a shared foundation that ensures cross-graph and cross-task interpretability and transferability.
\textbf{(c) Scalability.} Language-based IDs can be generated on the fly for any node, whether previously seen or not, and remain compatible with existing ones, avoiding token explosion.

\noindent \textbf{Multi-Instruction Fine-Tuning.} 
As shown in Figure~\ref{Fig_GIM}, we use multi-instruction fine-tuning to integrate diverse graphs and tasks~\citep{DBLP:journals/jmlr/ChungHLZTFL00BW24, DBLP:conf/iclr/WeiBZGYLDDL22}. 
We index nodes from the vocabulary and insert their language-based IDs into task-oriented templates to construct instructions $\mathcal{T}$:
\begin{equation}
    \begin{aligned}
        \mathcal{T} \leftarrow \text{Prompt}_{\text{template}}\left( T, G \right),
    \end{aligned}
\end{equation}
\noindent which include the language-based IDs of central nodes and their task-specific local structures from $G$ (e.g., for node classification or link prediction). 
These instructions are fully readable and coherent as they are entirely composed of natural language tokens. 
Thus, graph-centric tasks can be naturally reformulated into a unified text-to-text paradigm for LLM fine-tuning~\citep{DBLP:conf/acl/MishraKBH22}. 
Given a target sequence $Y$, the loss is computed as:
\begin{equation}
    \begin{aligned}
        \mathcal{L} &= - \sum_{s=1}^{|Y|} \log \text{Pr} \left (y_s \mid \mathcal{T}, Y_{<s} \right),
    \end{aligned}
\end{equation}
\noindent where \( \Pr\left( y_s \mid \mathcal{T}, Y_{<s} \right) \) denotes the probability of the \(s\)-th output token \(y_s\), conditioned on the instruction \(\mathcal{T}\) and previous tokens \(Y_{<s} = (y_1, y_2, \dots, y_{s-1})\). 
To support broad generalization, we fine-tune LLM backbones such as Flan-T5~\citep{DBLP:journals/jmlr/ChungHLZTFL00BW24} and Llama~\citep{DBLP:journals/corr/abs-2302-13971} to co-train a unified open-world knowledge model with cross-graph and cross-task generalization. 
As a result, we not only facilitate robust knowledge transfer across diverse unseen graphs and tasks (\S\ref{sec: cross-graph_cross-task_transfer}), but also exhibit adaptability to accommodate auxiliary pre-training tasks (\S\ref{sec: ood_pretraining_tasks}).

\noindent \textbf{Constrained Decoding via Prefix Tree Search.} 
To reduce LLM hallucination in generative tasks, we use constrained decoding with prefix tree search~\citep{DBLP:conf/iclr/CaoI0P21, DBLP:conf/sigir/TanXHGLZ24}. 
Taking generative link prediction as an example, we build a prefix tree from the language-based IDs of all candidate nodes, where each tree node is a natural language token and each root-to-leaf path denotes a valid node ID. 
During autoregressive generation, token selection is restricted to valid prefix-tree paths, ensuring predictions correspond to actual graph nodes and preventing hallucinated outputs. 
This effectiveness can be attributed to the discrete nature of language-based IDs, further highlighting the flexibility of our proposed graph vocabulary.

\begin{table*}[!tb]
\footnotesize
\vspace{-3mm}
\renewcommand{\arraystretch}{0.9}
\centering
    \caption{Evaluation results ($\%$) on node classification accuracy ($\uparrow$). We highlight the \color{darkred}{\textbf{first}}\color{black}{ and} \color{teal}{\textbf{second}} \color{black} best results, and report the statistical significance over the strongest baseline using a two-tailed $t$-test.}
\label{Node Classification}
    \setlength{\tabcolsep}{8pt}
\resizebox{\linewidth}{!}{
\begin{tabular}{c|ccccccc}
\toprule

Method           & Cora  & {Citeseer} & {PubMed} & {Ogbn-arxiv} & {History} & {Photo} & {WikiCS} \\ \midrule
{MLP}             & 62.29       & 64.42       & 62.88    & 62.07    &   64.62     &   61.21   &   68.41     \\ 
{GCN}             & 82.47       & 76.11       & 77.36    & 66.15    &   81.93     &   78.58   &   76.33     \\
{GAT}             & 82.92       & 77.30       & 74.36    & 65.29    &   82.85     &   82.38   &   78.21          \\
{GraphSAGE}            & 83.69       & 73.17       & 83.22    & 68.78    &   82.12     &   80.06   &   79.56          \\
{RevGNN}          & 86.90       & 77.34       & 82.16    & 70.43    &   \color{teal}\textbf{83.04}     &   83.24   &   81.22          \\
{SGFormer}        & 82.36       & 73.76       & 78.92    & 63.44        &      78.98           &      80.12           &        76.56          \\
{NodeFormer}      & 81.55       & 72.98       & 76.49    & 73.21        &      79.60           &      78.51           &      75.47            \\ \midrule
{BERT}            & 79.02       & 72.83       & 76.74    & 71.90        &      72.97        &     68.82            &     77.98             \\
{Sentence-BERT}    & 78.09       & 72.12       & 75.48    & 77.24        &      74.10           &      69.02           &      77.72            \\
{Llama3}    & 78.17       & 74.68       & 77.46    & 77.65        &      76.49           &      74.42           &      79.94            \\\midrule
{OFA}   & 75.72      & 71.58      & 75.26           & 74.68               &     81.43            &       84.46          &        78.02          \\
{LLaGA} & 81.25      & 68.80      & 86.54           & 76.05               &      82.55           &     {85.34}            &    80.74              \\
{ENGINE}          & 91.48         & 78.46           & 90.24          & 76.02        &  82.46               &      83.75           &  \color{teal}\textbf{81.56}                \\ 
{GraphPrompter}   & 80.26         & 73.61           &\color{darkred}{\textbf{94.80}}           & 75.61               &      79.42           &    80.04             &        80.98          \\
{GFT}             & 79.88         & 71.45           & 80.21          & 73.11        &  75.58               &      76.24           &  81.02                \\
{GraphText}       & 88.23         & 76.64           & 85.19          & 76.80        &  81.47               &      82.26           &  81.49                \\ \midrule
{PromptGFM (Flan-T5)}            & \color{teal}{\textbf{91.72}} & \color{teal}{\textbf{84.49}}    & 92.83  & \color{teal}{\textbf{80.58}}      &           82.33      &          \color{teal}\textbf{85.41}       &  81.49     \\
{PromptGFM (Llama3-8B)}            & \color{darkred}{\textbf{92.42}} & \color{darkred}{\textbf{85.32}}    & \color{teal}\textbf{94.65}  & \color{darkred}\textbf{83.78} &  \color{darkred}\textbf{86.72}      &          \color{darkred}\textbf{86.61}      &   \color{darkred}\textbf{84.66}   
 \\ \midrule
$p$-value      & 4e-3       & 2.73e-8       & 3.21e-7    & 4.48e-10        &     2.64e-6      &     3.62e-2         &     9.93e-4    \\

 \bottomrule
\end{tabular}
}
\vspace{-4mm}
\end{table*}

\begin{table}[!tbp]
    \footnotesize
    \centering
    \renewcommand{\arraystretch}{0.95}
    \caption{Accuracy ($\%$) on \underline{discriminative} link prediction.}
    \label{Discriminative Link Prediction}
    \setlength{\tabcolsep}{4pt}
        \begin{tabular}{c|cccc}
            \toprule
            \multirow{2}{*}{Method} &  \multicolumn{4}{c}{{Discriminative: Accuracy ($\uparrow$)}}  \\
             & Cora & Citeseer & Ogbn-arxiv & PubMed \\
            \midrule
            {GCN}           & 77.15      & 78.72      & 80.89           & 77.36     \\
            {GAT}           & 70.44      & 77.17      & 76.25           & 74.36      \\
            {GraphSAGE}          & 85.31      & 87.15      & 80.76           & 83.22      \\
            \midrule
            {GraphPrompter}       & 90.10      & 91.67      & 73.21           & 80.49   \\
            \midrule
            {Ours (Flan-T5)}  & \color{teal}\textbf{90.57}      & \color{teal}\textbf{92.03}      & \color{teal}\textbf{81.12}      & \color{teal}\textbf{87.64}         \\
            {Ours (Llama3)}     & \color{darkred}\textbf{91.68}      & \color{darkred}\textbf{93.46}      & \color{darkred}\textbf{84.27}           & \color{darkred}\textbf{88.12}    \\
            \bottomrule
        \end{tabular}
\end{table}

\subsection{Computational Complexity}
\label{Computational Complexity}
\textbf{Graph Understanding Module.} 
Although PromptGFM incurs an initial API cost during graph understanding, this cost is a one-time overhead and remains manageable with modern infrastructure.
Each graph is processed once to generate language-based IDs, which are then cached and reused throughout inference. When new graphs are introduced, existing IDs remain unchanged, and only the new graphs require ID generation. Subsequent LLM fine-tuning over the shared vocabulary then incorporates the features of these new graphs.


\noindent \textbf{Graph Inference Module.} 
With IDs cached, fine-tuning and decoding operate entirely over ID tokens. 
At inference time, each decoding step costs $O(\bar{m})$, where $\bar{m}$ is the average length of a language-based ID. 
With prefix tree search constraining the candidate space to the $|V|$ nodes of the graph-specific vocabulary, the overall decoding complexity is bounded by $O(|V| \times \bar{m})$. 
Thus, these properties keep the computational overhead manageable for large-scale graphs. An empirical efficiency study is provided in \S\ref{exp: efficiency_analysis}.

\section{Experiments}
\label{sec: exp}

In this section, we conduct extensive experiments to address these research questions (RQs): 
\textbf{RQ1:} How does PromptGFM perform on node classification and link prediction?
\textbf{RQ2:} How does PromptGFM transfer across diverse graphs and tasks as a general-purpose GFM?
\textbf{RQ3:} How does PromptGFM adapt to more auxiliary pre-training tasks?
\textbf{RQ4:} How does each module contribute to the overall performance?
\textbf{RQ5:} What is the computational efficiency in our graph understanding and inference modules?
We evaluate PromptGFM on seven TAG datasets from computer science, e-commerce, and biomedical domains. We compare against graph-agnostic, GNN-based, LM-only, and GNN-LLM integration baselines, and report node classification and link prediction performance under consistent evaluation protocols. Dataset statistics are summarized in Appendix~\ref{app:experimental-setup}, baseline details in Appendix~\ref{app: baselines}, and reproduction settings in Appendix~\ref{app: implementation_details}.

\begin{table}[!tbp]
    \footnotesize
    \centering
    \renewcommand{\arraystretch}{0.95}
    \setlength{\tabcolsep}{10pt}
    \caption{HR@1 ($\%$) on \underline{generative} link prediction.}
    \label{Generative Link Prediction}
        \begin{tabular}{c|ccc}
            \toprule
            \multirow{2}{*}{Method} &  \multicolumn{3}{c}{{Generative: HR@1 ($\uparrow$)}}  \\
             & Cora & Citeseer & PubMed \\
            \midrule
            {GCN}         &   5.95    &     6.82      &     0.51     \\
            {GAT}         &   2.22    &     3.59      &     0.28       \\
            {GraphSAGE}        &   6.59    &     8.73      &     0.45          \\
            \midrule
            {InstructGLM} &   6.65    &     7.13      &     0.58          \\
            {KGT5-context} &  8.03    &     8.41      &     1.17          \\
            \midrule
            PromptGFM   &      \color{darkred}\textbf{8.21}      &    \color{darkred} \textbf{8.90}    &   \color{darkred}\textbf{1.21}         \\
            \bottomrule
        \end{tabular}
\end{table}

\subsection{Main Performance Comparison (RQ1)}  

\textbf{Main Results.} 
We train each model from scratch on a single graph and report NC and LP accuracy in Tables~\ref{Node Classification} and~\ref{Discriminative Link Prediction}. 
PromptGFM substantially outperforms state-of-the-art baselines, leading to the following observations. 
(a) Graph-based models consistently outperform graph-agnostic ones, confirming the importance of structural information. 
(b) PromptGFM surpasses OFA and ENGINE on node classification, as well as GraphPrompter on link prediction, suggesting that decoupled GNN-for-LLM and LLM-for-GNN architectures have limited graph-text alignment. 
(c) PromptGFM also outperforms LLaGA, indicating that heuristic prompting alone is insufficient to capture higher-order dependencies without a true GNN mechanism, whereas our prompt-based GNN performs explicit message passing. 
(d) PromptGFM also surpasses the vocabulary-based GFT and the text-space GraphText, which respectively lack a language-grounded vocabulary and an iterative aggregation-update process, showing that the two components of our design are complementary. 
(e) Notably, the gains also hold on History and Photo, whose co-purchase and co-view edges relate items behaviorally rather than lexically, indicating that PromptGFM selectively integrates informative neighbor signals and generalizes beyond lexically homophilous graphs.

\begin{table}[!tbp]
    \footnotesize
    \centering
    \renewcommand{\arraystretch}{0.95}
    \setlength{\tabcolsep}{5pt}
    \caption{Intra-domain cross-graph transferability to \underline{Citeseer} within the \underline{computer science} domain.}
    \label{tab: intra_domain}
    \begin{tabular}{cc|cc|cc}
        \toprule
        \multicolumn{2}{c|}{Source} & \multicolumn{2}{c|}{Pre-training} & \multicolumn{2}{c}{Co-training} \\
        Cora      & Arxiv      & Acc($\uparrow$)   & M-F1($\uparrow$)   & Acc($\uparrow$)   & M-F1($\uparrow$)   \\
        \midrule
        {\xmark}   & {\xmark}   & 27.64  & 17.10  & 84.49  & 82.31  \\ \midrule
        {\cmark}   & {\xmark}   & 51.63  & 45.10  & 84.96  & 83.22  \\
        {\xmark}   & {\cmark}   & \color{teal}\textbf{60.34} & \color{teal}\textbf{54.81} & \color{teal}\textbf{85.45} & \color{teal}\textbf{83.91} \\ \midrule
        {\cmark}   & {\cmark}   & \color{darkred}\textbf{61.25} & \color{darkred}\textbf{55.66} & \color{darkred}\textbf{86.77} & \color{darkred}\textbf{84.24} \\
        \bottomrule
    \end{tabular}
\end{table}

\begin{table}[!tbp]
    \footnotesize
    \centering
    \renewcommand{\arraystretch}{0.95}
    \setlength{\tabcolsep}{4pt}
    \caption{Inter-domain cross-graph transferability from \underline{e-commerce} to \underline{biomedical} domain.}
    \label{tab: inter_domain_1}
    \begin{tabular}{cc|cc|cc}
        \toprule
        \multicolumn{2}{c|}{Source} & \multicolumn{2}{c|}{Pre-training} & \multicolumn{2}{c}{Co-training} \\
        History   & Photo      & Acc($\uparrow$)   & M-F1($\uparrow$)   & Acc($\uparrow$)   & M-F1($\uparrow$)   \\
        \midrule
        {\xmark}   & {\xmark}   & 39.12  & 39.84  & \color{darkred}\textbf{90.67} & \color{darkred}\textbf{91.82} \\ \midrule
        {\cmark}   & {\xmark}   & \color{darkred}\textbf{46.24} & \color{darkred}\textbf{47.41} & \color{teal}\textbf{88.52}    & \color{teal}\textbf{89.39}    \\
        {\xmark}   & {\cmark}   & 44.21  & 44.93  & 85.34  & 85.57  \\ \midrule
        {\cmark}   & {\cmark}   & \color{teal}\textbf{45.18} & \color{teal}\textbf{45.65} & 86.22  & 87.03  \\
        \bottomrule
    \end{tabular}
\end{table}

\noindent \textbf{Generative Link Prediction.} 
\label{sec: generative_link_prediction}
Following the transductive setting, we partition the graph by links, construct the input graph from training edges, and predict missing connections for test nodes with existing nodes as candidates. 
As shown in Table~\ref{Generative Link Prediction}, PromptGFM consistently outperforms both traditional GNNs and generative sequence-to-sequence baselines~\citep{DBLP:conf/eacl/YeZWXZ24, kochsiek2023friendly} on generative link prediction.
Existing GNN-LLM methods fail on this task because their OOV token embeddings cannot be directly mapped from LLM outputs to specific nodes, often causing hallucinated predictions. 
In contrast, PromptGFM represents nodes as finite token sequences and uses prefix-tree constrained decoding to regulate LLM outputs. 
This highlights the importance of graph vocabulary and the flexibility of PromptGFM in generative scenarios.

\subsection{Cross-graph and Cross-task Transfer (RQ2)}
\label{sec: cross-graph_cross-task_transfer}

We evaluate cross-graph transferability under both \textbf{intra-domain} and \textbf{inter-domain} scenarios. 
Following prior work~\citep{DBLP:conf/nips/ChenMLSLJFTPLT24}, we consider two settings: 
\textbf{(a) Pre-training}, where models are evaluated on unseen datasets not used during training, and 
\textbf{(b) Co-training}, where part of the target dataset is jointly trained with auxiliary datasets. 
These two dimensions define four transfer paradigms for a comprehensive assessment of cross-graph transferability.

\begin{table}[!tbp]
    \footnotesize
    \centering
    \renewcommand{\arraystretch}{0.95}
    \setlength{\tabcolsep}{2pt}
    \caption{Inter-domain cross-graph transferability from \underline{computer science} to \underline{biomedical} domain.}
    \label{tab: inter_domain_2}
    \begin{tabular}{ccc|cc|cc}
        \toprule
        \multicolumn{3}{c|}{Source}   & \multicolumn{2}{c|}{Pre-training} & \multicolumn{2}{c}{Co-training} \\
        Cora        & Citeseer      & Arxiv         & Acc($\uparrow$)        & M-F1($\uparrow$)        & Acc($\uparrow$)        & M-F1($\uparrow$)        \\
        \midrule
        {\xmark}        & {\xmark}      & {\xmark}        & 39.12         & 39.84           & \color{darkred}\textbf{90.67}        & \color{darkred}\textbf{91.82}          \\ \midrule
        {\cmark}        & {\xmark}      & {\xmark}        & 51.76         & 52.84           & \color{teal}\textbf{86.34}       & \color{teal}\textbf{87.29}         \\
        {\xmark}        & {\cmark}      & {\xmark}        & 40.12         & 42.38           & 85.21           & 86.12           \\
        {\xmark}        & {\xmark}      & {\cmark}        & \color{darkred}\textbf{60.21}        & \color{darkred}\textbf{62.02}           & 82.17           & 86.01           \\ \midrule
        {\cmark}        & {\cmark}      & {\xmark}        & 50.17         & 51.74           & 84.39         & 81.94           \\
        {\cmark}        & {\xmark}      & {\cmark}        & \color{teal}\textbf{57.28}       & \color{teal}\textbf{59.71}           & 81.32         & 83.14           \\
        {\xmark}        & {\cmark}      & {\cmark}        & 55.34         & 57.11           & 82.13         & 83.10           \\ \midrule
        {\cmark}        & {\cmark}      & {\cmark}        & 53.07         & 54.90          & 80.43       & 80.79           \\
        \bottomrule
    \end{tabular}
\end{table}

\begin{table}[!tbp]
    \footnotesize
    \centering
    \renewcommand{\arraystretch}{0.95}
    \caption{Cross-task generalization between node classification \underline{(NC)} and link prediction \underline{LP}. \textit{ST-X} refers to supervised training on the task X from scratch.}
    \label{Cross-task performance from link prediction to node classification.}
    \setlength{\tabcolsep}{8pt}
    \begin{tabular}{c|c|c|c|c}
        \toprule
        Task & Setting & Cora & Citeseer & PubMed \\ 
        \midrule
        \multirow{3}{*}{NC} 
        & \textit{zero-shot}      & 18.54 & 27.64 & 39.12 \\
        & \textit{LP$\rightarrow$NC} & 60.74 & 50.12 & 57.42 \\
        & \textit{ST-NC}          & \color{darkred}\textbf{91.72} & \color{darkred}\textbf{84.49} & \color{darkred}\textbf{92.83} \\
        \midrule 
        \multirow{3}{*}{LP} 
        & \textit{zero-shot}       & 67.28 & 69.22 & 73.59 \\
        & \textit{NC$\rightarrow$LP} & 78.16 & 79.23 & 84.11 \\
        & \textit{ST-LP}            & \color{darkred}\textbf{91.68} & \color{darkred}\textbf{93.46} & \color{darkred}\textbf{88.12} \\
        \bottomrule
    \end{tabular}
\end{table}
\noindent \textbf{Intra-domain Cross-graph Transferability.}
As shown in Table~\ref{tab: intra_domain}, all settings outperform direct inference with an off-the-shelf LLM. 
Incorporating Arxiv brings larger gains than Cora, suggesting that larger and richer datasets improve transferability more effectively. 
In the \textit{pre-training} setting, the substantial gains over direct inference indicate strong in-domain zero-shot transferability. 
The \textit{co-training} setting shows a similar but smaller trend, further validating the effectiveness of intra-domain knowledge transfer. 
These results highlight the value of collecting more relevant graph data to build more knowledgeable and transferable GFMs.

\noindent \textbf{Inter-domain Cross-graph Transferability.} 
We report transfer results from multiple domains to the biomedical domain in Tables~\ref{tab: inter_domain_1} and~\ref{tab: inter_domain_2}. 
In the \textit{pre-training} setting, adding a single external graph consistently improves performance, whereas multiple source graphs may slightly degrade target-domain performance. 
More importantly, in the supervised \textit{co-training} setting, performance can deteriorate as more cross-domain data is introduced.
This is mainly due to domain shifts, including structural differences, semantic drift, and mismatched label spaces~\citep{DBLP:conf/nips/ChenMLSLJFTPLT24, DBLP:conf/kdd/ZhaoCS0L24}. 
Nonetheless, PromptGFM adopts a unified language-based vocabulary and pure language interface, which in principle enables natural cross-domain transferability and avoids token collision.
We discuss this community-wide challenge and potential solutions in Appendix~\ref{sec:adaptive_prompting}.


%

\noindent \textbf{Cross-task Transferability.}
We present transfer performance between LP and NC in Table~\ref{Cross-task performance from link prediction to node classification.}. 
As expected, \textit{LP$\rightarrow$NC} generally surpasses \textit{zero-shot}, showing successful transfer to unseen tasks, but remains far below \textit{ST-NC}, indicating that cross-task transfer is harder than cross-graph generalization. 
A similar trend appears for LP: \textit{NC$\rightarrow$LP} outperforms zero-shot but still trails \textit{ST-LP}, suggesting that PromptGFM transfers task-agnostic structural knowledge while task-specific supervision remains important. 
Overall, these results demonstrate our cross-graph and cross-task generalization capability, establishing a task-adaptive GFM.

\begin{table}[tbp]
    \centering
    \caption{The impact of incorporating Node Discrimination \underline{(+ND)} and Link Discrimination \underline{(+LD)} as pre-training tasks on NC performance.}
    \label{tab:pretraining_ablation}
    \footnotesize
    \renewcommand{\arraystretch}{1}
    \resizebox{\linewidth}{!}{
    \begin{tabular}{l cc cc}
        \toprule
        {Dataset} & {+ND} & {+LD} & {Accuracy (\%)} & {Macro-F1 (\%)} \\
        \midrule
        \multirow{4}{*}{Citeseer} 
        & \xmark & \xmark & 84.49 & 82.31 \\
        & \cmark & \xmark & 85.67 ($\uparrow$1.18) & 83.22 ($\uparrow$0.91) \\
        & \xmark & \cmark & 84.73 ($\uparrow$0.24) & 82.65 ($\uparrow$0.34) \\
        & \cmark & \cmark & \color{darkred}\textbf{86.02} ($\uparrow$1.53) & \color{darkred}\textbf{83.77} ($\uparrow$1.46) \\
        \midrule
        \multirow{4}{*}{PubMed}
        & \xmark & \xmark & 90.67 & 91.82 \\
        & \cmark & \xmark & 92.53 ($\uparrow$1.86) & 92.84 ($\uparrow$1.02) \\
        & \xmark & \cmark & 91.12 ($\uparrow$0.45) & 92.03 ($\uparrow$0.21) \\
        & \cmark & \cmark & \color{darkred}\textbf{93.06} ($\uparrow$2.39) & \color{darkred}\textbf{93.27} ($\uparrow$1.45) \\
        \bottomrule
    \end{tabular}
    }
\end{table}

\subsection{Exploration Studies (RQ3\textasciitilde5)}
\label{Exploration Studies}

\begin{figure}[!tbp]
    \centering
    \caption{Ablation studies on node classification.}
    \includegraphics[width=0.96\linewidth]{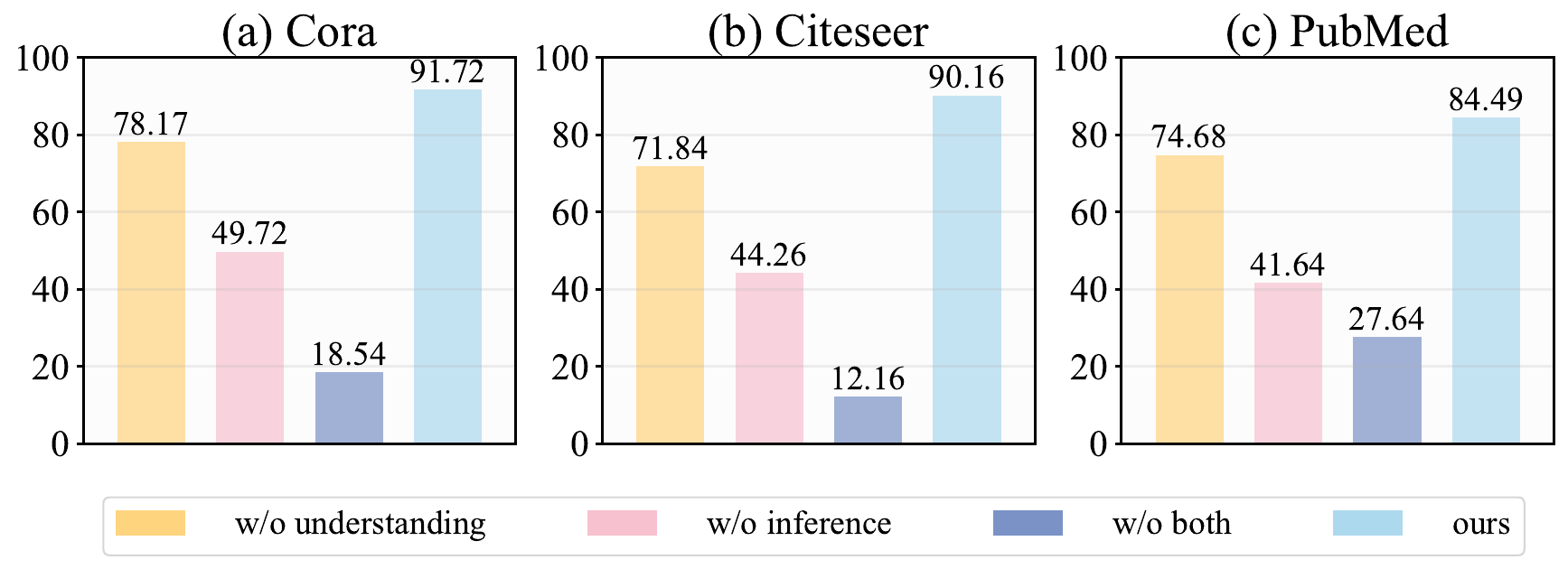}
    \label{Ablation Studies}
\end{figure}

\label{sec: ood_pretraining_tasks}

\textbf{Adding Auxiliary Pre-training Tasks (RQ3).} 
A robust foundation model should support diverse pre-training tasks to improve downstream generalization. 
Beyond NC and LP, we introduce two additional tasks: 
\textbf{(a) Node Discrimination (ND):} \textit{Given three nodes [a], [b], [c], exactly one belongs to a different category. Which one is it?} 
and \textbf{(b) Link Discrimination (LD):} \textit{Given a central node [X], which candidate node [a], [b], [c] is not connected to [X]?}, 
where bracketed symbols denote nodes' language-based IDs. 
We incorporate ND and LD individually or jointly into multi-task fine-tuning and evaluate their effects on NC under identical settings.
As shown in Table~\ref{tab:pretraining_ablation}, adding either ND or LD improves both accuracy and M-F1, while combining them achieves the best performance. 
These results show that auxiliary tasks enrich node representations and improve target-task performance. 
Thus, PromptGFM goes beyond memorizing task patterns and leverages its unified language interface to exhibit stronger cross-task generalization.

\noindent \textbf{Ablation Studies (RQ4).} 
We construct three ablated variants: \textbf{(a) w/o understanding}, which removes the prompt-based GNN and instead uses the summarized central-node text from the initialization prompt as the language-based ID; \textbf{(b) w/o inference}, which removes multi-instruction fine-tuning; and \textbf{(c) w/o both}, which directly prompts LLMs with raw textual attributes.
Figure~\ref{Ablation Studies} shows NC accuracy, where the full model outperforms all variants. 
The drop in \textit{w/o understanding} highlights the importance of replicating the GNN workflow to preserve semantic and structural signals. 
The larger decline in \textit{w/o inference} emphasizes the necessity of LLM fine-tuning for incorporating domain-specific knowledge. 
Finally, \textit{w/o both} performs worst, confirming the complementarity between graph understanding and inference.

\noindent \textbf{Controlled Neighbor Study.}
To further isolate the contribution of the graph understanding module, we construct three controlled variants that replace the sampled neighbors while keeping the same backbone, prompt template, token budget, and padding strategy: \textbf{(a) central text only}, which removes neighbor context; \textbf{(b) semantic non-neighbors}, which uses Sentence-BERT-nearest disconnected nodes; and \textbf{(c) degree-preserving rewired}, which samples neighbors from a degree-preserving rewired graph. Table~\ref{tab: controlled_neighbor} reports NC accuracy and M-F1, where true neighbors consistently outperform all variants. The drop in \textit{central text only} shows the benefit of neighbor context beyond the central node. The remaining gap in \textit{semantic non-neighbors} indicates that semantic relatedness cannot substitute for graph topology. Finally, \textit{degree-preserving rewired} performs worst, even below \textit{central text only} despite the same amount of neighbor text, confirming that our prompt-based GNN relies on the given topology.

\begin{table}[!tb]
    \centering
    \vspace{-2mm}
    \caption{Controlled neighbor study on node classification, reporting accuracy and Macro-F1 (\%) results.}
    \label{tab: controlled_neighbor}
    \footnotesize
    \setlength{\tabcolsep}{3pt}
    \renewcommand{\arraystretch}{1.15}
    \begin{tabular}{l|cc|cc}
        \toprule
        \multirow{2}{*}{Variant} & \multicolumn{2}{c|}{Cora} & \multicolumn{2}{c}{Citeseer} \\
         & Acc & M-F1 & Acc & M-F1 \\
        \midrule
        Central text only & 78.17 & 75.44 & 74.68 & 71.19 \\
        Semantic non-neighbors & 82.54 & 81.27 & 75.24 & 72.36 \\
        Degree-preserving rewired & 61.13 & 58.71 & 62.48 & 60.04 \\
        \midrule
        PromptGFM (true neighbors) & \color{darkred}\textbf{91.72} & \color{darkred}\textbf{90.06} & \color{darkred}\textbf{84.49} & \color{darkred}\textbf{80.13} \\
        \bottomrule
    \end{tabular}
\end{table}

\noindent \textbf{Efficiency Analysis (RQ5).}
\label{exp: efficiency_analysis}
We evaluate the computational and monetary cost of PromptGFM, as summarized in Table~\ref{tab: efficiency}. In the graph understanding module, the API and token cost grow with the graph size, but this constitutes a one-time, graph-specific overhead that can be cached and reused throughout the inference pipeline (denoted as ``n/a''). On the largest graph, Ogbn-arxiv, this one-time graph understanding stage costs \$2.69 and 17.91M tokens (about 176 minutes and 1.41 kWh), which is amortized by ID caching. During inference, as more graphs are introduced and the overall vocabulary expands, we observe that time and energy consumption increase in an approximately linear manner. These results align with our discussion in \S\ref{Computational Complexity} and demonstrate PromptGFM's efficiency as a scalable GFM.



\section{Related Works}
\label{sec: related_works}

\noindent \textbf{GNN-LLM Integration.} 
Motivated by the potential of LLMs for graph learning, existing GNN-LLM integration methods for TAGs fall into three paradigms: 
\textbf{(a) GNN for LLM}, where GNNs or graph transformers act as structure tokenizers to enhance LLM understanding~\citep{DBLP:conf/sigir/Tang00SSCY024, DBLP:journals/corr/abs-2310-05845, DBLP:conf/www/0011H0C24}; 
\textbf{(b) LLM for GNN}, where LLMs generate labels or features to supervise GNN training~\citep{DBLP:conf/iclr/ChenMWH0Z0T24, DBLP:conf/iclr/0057FKLT0Z24, DBLP:conf/ijcai/ZhuWST24, DBLP:conf/iclr/HeB0PLH24, DBLP:journals/corr/abs-2402-15183}; and 
\textbf{(c) LLM as GNN}, where LLMs consume structured prompts but lack an intrinsic GNN mechanism~\citep{DBLP:conf/icml/Chen0JSW24, DBLP:conf/eacl/YeZWXZ24, DBLP:conf/acl/WangWH00M24, DBLP:journals/corr/abs-2310-01089}. 
All paradigms rely on decoupled two-stage alignment and share a common co-training challenge, motivating our unified LLM-as-GNN paradigm.

\begin{table}[tbp]
    \centering
    \vspace{-2mm}
    \setlength{\tabcolsep}{3pt}
    \renewcommand\arraystretch{1}
    \caption{Efficiency analysis in terms of API cost (USD), number of tokens (M), inference time (min), and power consumption (kWh).}
    \vspace{-1.5mm}
    \label{tab: efficiency}
    \resizebox{\linewidth}{!}{
    \begin{tabular}{ccc|c|c|c|c}
        \toprule
        \multicolumn{3}{c|}{Source} & \multicolumn{2}{c|}{Understanding} & \multicolumn{2}{c}{Inference} \\
        Cora & Citeseer & Photo & API Cost & \# Tokens & Time & Power \\
        \midrule
        {\cmark} & {\xmark} & {\xmark} & 0.188 & 1.25 & 16.47 & 0.07 \\
        {\xmark} & {\cmark} & {\xmark} & 0.189 & 1.29 & 21.41 & 0.08 \\
        {\xmark} & {\xmark} & {\cmark} & 1.213 & 8.08 & 83.67 & 0.34 \\
        {\cmark} & {\cmark} & {\xmark} & n/a & n/a & 35.31 & 0.16 \\
        {\cmark} & {\cmark} & {\cmark} & n/a & n/a & 117.56 & 1.04 \\
        \bottomrule
    \end{tabular}
    }
\end{table}

\noindent \textbf{Graph-Text Alignment.} 
Existing methods align graph and text representations through graph-encoder prefixes~\citep{DBLP:journals/corr/abs-2309-02848, DBLP:conf/www/HuangHYBTCZ24}, contrastive learning~\citep{DBLP:conf/sigir/Tang00SSCY024, DBLP:conf/www/ZhuSWLWPHT25, DBLP:conf/emnlp/0001DL23, DBLP:journals/corr/abs-2305-14321}, iterative training~\citep{DBLP:conf/ijcai/ZhuWST24, DBLP:conf/iclr/0002QLYL00023}, knowledge distillation~\citep{DBLP:conf/pkdd/MavromatisIWZAMZFK23}, or structure-aware self-supervised learning~\citep{zhang2025taga, liu2025sstag, yin2025mmsa}.
However, they still suffer from a modality gap that limits transferability. 
Since TAGs already contain textual information, we shift graph-text alignment entirely into text space in this work.

\noindent \textbf{Graph Foundation Models.} GFMs aim to transfer across diverse datasets and tasks, with a central challenge being the identification of a graph vocabulary of transferable units~\citep{DBLP:conf/www/0008HZW0Y25, DBLP:conf/wsdm/FangFDLT25, DBLP:journals/corr/abs-2310-11829}. Existing approaches rely on domain-specific vocabularies: GraphGPT~\citep{DBLP:conf/sigir/Tang00SSCY024} assigns unique IDs to each dataset, MoleBERT~\citep{DBLP:conf/iclr/XiaZHG0LLL23} encodes molecular structures, and GFT~\citep{DBLP:conf/nips/WangZCZ024} quantizes computation trees into a discrete tree vocabulary. However, such vocabularies do not readily transfer across domains. While prior work has represented and reasoned over graphs in natural language~\citep{DBLP:conf/iclr/FatemiHP24, DBLP:conf/nips/WangFHTHT23}, natural language has yet to be explored as a transferable graph vocabulary.
\vspace{-0.05cm}
\section{Conclusion}
\vspace{-0.15cm}
We present PromptGFM, a graph foundation model for text-attributed graphs using graph vocabulary learning. By replicating the GNN workflow in text space, PromptGFM decouples refined textual node representations into a unified vocabulary, preserving graph–text semantic fidelity while avoiding out-of-vocabulary issues. This vocabulary improves compatibility and scalability, enabling effective LLM fine-tuning with readable instructions and stronger transfer. Experiments show superior performance, cross-graph and cross-task generalization, and adaptability to auxiliary pre-training tasks, highlighting the potential of LLMs as GNNs and new directions for graph foundation models.

\section*{Limitations}
\addcontentsline{toc}{section}{Limitations}

We discuss several aspects of the current design that warrant further exploration.

\noindent \textbf{(1) Closed-source LLM for Node Initialization.} Our graph understanding module uses GPT-4o mini to generate initial node summaries. Because the resulting language-based IDs are cached and reused, the associated API cost is bounded and amortized over subsequent use. Nevertheless, replacing the closed-source model with capable open-source alternatives could further improve the accessibility and reproducibility of our framework.

\noindent  \textbf{(2) Prompt Length for Dense Graphs.} For graphs with highly skewed degree distributions, one-hop prompts for high-degree nodes can become lengthy. In our experiments, neighbor sampling and a fixed neighbor cap, as described in Appendix~\ref{Prompt Length Limitation Discussion}, effectively bound the prompt length. For substantially larger or denser graphs, additional techniques such as hierarchical neighborhood summarization could further improve scalability.

\noindent \textbf{(3) Dynamic Graphs.} Our current evaluation mainly focuses on static graphs, where language-based IDs are generated once and then reused. For dynamic graphs, we introduce a localized update strategy that regenerates representations only for affected nodes, with preliminary results reported in Appendix~\ref{sec: dynamic_graphs}. A more systematic investigation of large-scale streaming graphs and frequent updates remains future work.

\noindent  \textbf{(4) Cross-domain Transferability.} Our language-based vocabulary naturally supports knowledge sharing across domains. However, the adaptive prompting experiments in Appendix~\ref{sec:adaptive_prompting} show that transfer between highly different source and target domains remains challenging, especially under co-training. We view robust cross-domain alignment as an open research direction for graph foundation models more broadly, and adaptive prompting offers a promising step toward it.

\section*{Ethical Considerations}
\addcontentsline{toc}{section}{Ethical Considerations}

\noindent \textbf{Data.} All datasets are publicly released academic benchmarks (Cora, Citeseer, PubMed, Ogbn-arxiv, WikiCS, Amazon Photo/History). They contain no personally identifiable information. We follow the licenses and intended uses of each dataset.

\noindent \noindent \textbf{Models.} We build on public LLM backbones (Flan-T5, Llama3) and call OpenAI's GPT-4o mini for the one-time graph understanding step. Outputs of LLMs may reflect biases inherited from their pre-training corpora, and the language-based IDs we derive inherit this risk. For sensitive deployments (e.g., recommender systems, biomedical decision support), downstream calibration and human review are warranted.

\noindent \textbf{Compute and Energy.} We report API cost, token count, inference time, and power consumption in Table~\ref{tab: efficiency}. Because language-based IDs are cached and reused, the dominant cost is one-time per graph, which limits the marginal carbon footprint of running PromptGFM at deployment.


\noindent \textbf{AI Use Disclosure.} AI assistants were used solely for language polishing, including grammar correction and readability improvement. All research ideas, methods, experiments, analyses, and conclusions are the authors' own, and the authors take full responsibility for the content.

\bibliography{software}

\clearpage
\onecolumn
\tableofcontents
\clearpage
\twocolumn
\appendix
\setlength{\abovecaptionskip}{10pt}
\setlength{\belowcaptionskip}{0pt}
\setlength{\textfloatsep}{20pt plus 2pt minus 4pt}
\setlength{\floatsep}{12pt plus 2pt minus 2pt}
\setlength{\intextsep}{12pt plus 2pt minus 2pt}
\setlength{\abovedisplayskip}{10pt plus 2pt minus 5pt}
\setlength{\belowdisplayskip}{10pt plus 2pt minus 5pt}

\section{Datasets}
\label{app:experimental-setup}
\label{app: datasets}

We use seven public benchmarking datasets from three domains to evaluate PromptGFM, with statistics summarized in Table~\ref{tab: data statistics}. Detailed descriptions of each dataset are provided below.

\subsection{Computer Science Datasets}

\noindent $\bullet$ \textbf{Cora}~\citep{DBLP:journals/ir/McCallumNRS00}. The dataset includes a citation network consisting of 2,708 scientific publications in the field of machine learning, categorized into 7 classes based on their research topics. Nodes represent individual papers, and edges denote citation links between them, totaling 10,858 connections. Each paper is associated with its title and abstract.

\noindent $\bullet$ \textbf{Citeseer}~\citep{DBLP:conf/dl/GilesBL98}. This dataset comprises a citation network of 3,327 scientific publications, categorized into 6 classes, including Agents, Artificial Intelligence, Database, Information Retrieval, Machine Learning, and Human-Computer Interaction. Nodes correspond to documents, and edges represent citation relationships between them, amounting to 9,464 links.

\noindent $\bullet$ \textbf{Ogbn-arxiv}~\citep{DBLP:conf/nips/HuFZDRLCL20}. The ogbn-arxiv dataset is part of the Open Graph Benchmark and consists of a directed citation graph of 169,343 arXiv papers, categorized into 40 subject areas. Nodes represent individual papers, and edges indicate citation relationships, totaling over 2.3 million connections. Each paper includes textual data from its title and abstract.

\noindent $\bullet$ \textbf{WikiCS}~\citep{DBLP:journals/corr/abs-2007-02901}. The WikiCS dataset is a benchmark dataset derived from Wikipedia, designed for evaluating GNNs. It comprises a citation network with 11,701 nodes representing computer science articles and 431,726 edges corresponding to hyperlinks between them. The dataset features 10 classes corresponding to branches of computer science, with very high connectivity.

\subsection{E-commerce Datasets}

\noindent $\bullet$ \textbf{History}~\citep{DBLP:conf/nips/YanLLY0ZYZHSDZ023}. The History dataset, extracted from the Amazon-Books dataset, includes items categorized under the second-level label ``History''. It comprises 41,551 nodes and 503,180 edges, where each node represents a book, and edges indicate frequent co-purchases or co-views between books. This dataset incorporates the title and description of each book as text attributes of the node. The classification task involves assigning books to 12 distinct categories.

\noindent $\bullet$ \textbf{Photo}~\citep{DBLP:conf/nips/YanLLY0ZYZHSDZ023}. The Photo dataset, derived from the Amazon-Electronics dataset, consists of 48,362 nodes and 873,782 edges, forming a dense network that reflects user purchasing behavior. Each node represents an electronic product, while edges indicate frequent co-purchases or co-views between items. For textual attributes, user reviews are incorporated, prioritizing the most upvoted review when available; otherwise, a randomly selected review is used. The classification task involves categorizing electronic products into 12 distinct categories.

\subsection{Biomedical Datasets}

\noindent $\bullet$ \textbf{PubMed}~\citep{DBLP:journals/aim/SenNBGGE08}. PubMed is a citation network of 19,717 scientific publications from the PubMed database pertaining to diabetes, classified into 3 classes: experimental induced diabetes, type 1 diabetes, and type 2 diabetes. Nodes are research papers, and edges signify citation links, amounting to 88,648 connections. This dataset is used for large-scale graph representation learning and evaluating algorithms in the biomedical domain.

\begin{table*}[!tb]
\centering
\caption{Statistics of seven benchmarking datasets from three domains.}
\label{tab: data statistics}
\footnotesize
\setlength{\tabcolsep}{18pt}
\renewcommand{\arraystretch}{1.15}
\begin{tabular}{ccccc}
\toprule
\textbf{Dataset} & \textbf{Domain} & \textbf{\#Nodes} & \textbf{\#Edges} & \textbf{\#Classes} \\ \midrule
Cora        & Computer Science & 2708   & 10858   & 7 \\
Citeseer    & Computer Science & 3327   & 9464    & 6 \\
Ogbn-arxiv  & Computer Science & 169343 & 2332486 & 40 \\
WikiCS      & Computer Science & 11701  & 431726  & 10 \\
Photo       & E-commerce       & 48362  & 873782  & 12 \\
History     & E-commerce       & 41551  & 503180  & 12 \\
PubMed      & Biomedical       & 19717  & 88648   & 3 \\
\bottomrule
\end{tabular}
\end{table*}

\section{Baselines}
\label{app: baselines}

We compare PromptGFM against a comprehensive set of baselines, organized into six categories.

\subsection{Graph-agnostic Methods}

\noindent $\bullet$ \textbf{MLP.} This method adopts a multi-layer perceptron to learn low-dimensional embeddings for each node without using graph structure.

\subsection{GNN-based Methods}

\noindent $\bullet$ \textbf{GCN}~\citep{DBLP:conf/iclr/KipfW17}. This model generalizes convolutional operations to graphs for semi-supervised learning.

\noindent $\bullet$ \textbf{GAT}~\citep{DBLP:journals/corr/abs-1710-10903}. This method integrates attention mechanisms to weigh neighbor features during aggregation.

\noindent $\bullet$ \textbf{GraphSAGE}~\citep{DBLP:conf/nips/HamiltonYL17}. GraphSAGE is an inductive framework that enables generalization to unseen data via neighborhood sampling and aggregation.

\noindent $\bullet$ \textbf{RevGNN}~\citep{DBLP:conf/icml/Li0GK21}. This method scales GNNs to over 1000 layers using reversible residual connections that bound memory consumption.

\noindent $\bullet$ \textbf{SGFormer}~\citep{DBLP:conf/nips/WuZYZNJBY23}. This work introduces a simplified graph transformer that captures both local and global graph structure with a single-layer global attention.

\noindent $\bullet$ \textbf{NodeFormer}~\citep{DBLP:conf/nips/WuZLWY22}. This framework presents a scalable graph transformer that utilizes a kernelized message-passing scheme, enabling efficient training on large-scale graphs.

\subsection{LM-only Methods}

\noindent $\bullet$ \textbf{BERT}~\citep{DBLP:conf/naacl/DevlinCLT19}. This language model learns bidirectional context through pre-training with masked language modeling and next sentence prediction.

\noindent $\bullet$ \textbf{Sentence-BERT}~\citep{DBLP:conf/emnlp/ReimersG19}. This model is a modification of BERT that uses a siamese network to generate fixed-size sentence embeddings.

\noindent $\bullet$ \textbf{Llama3-8B}~\citep{grattafiori2024llama3}. This is an open-weight, decoder-only Transformer with 8 billion parameters that achieves strong performance on a wide range of benchmarks, particularly in reasoning, multilingual tasks, and instruction following.

\subsection{GNN-LLM Integration Methods}

\noindent $\bullet$ \textbf{OFA}~\citep{DBLP:conf/iclr/0057FKLT0Z24}. This framework handles various graph classification tasks across different domains using a single model.

\noindent $\bullet$ \textbf{LLaGA}~\citep{DBLP:conf/icml/Chen0JSW24}. This model integrates LLMs for graph data by transforming nodes into structure-aware sequences and mapping them into the token embedding space with a projector.

\noindent $\bullet$ \textbf{GraphPrompter}~\citep{DBLP:conf/www/0011H0C24}. This framework aligns graphs with LLMs via soft prompts, including a GNN to capture structure and an LLM to interpret node text, combining semantic and structural insights through prompt tuning.

\noindent $\bullet$ \textbf{ENGINE}~\citep{DBLP:conf/ijcai/ZhuWST24}. This method presents a parameter- and memory-efficient fine-tuning approach that combines a tunable GNN-based side structure alongside each LLM layer.

\noindent $\bullet$ \textbf{GraphText}~\citep{DBLP:journals/corr/abs-2310-01089}. This framework converts graphs into hierarchical graph-syntax trees and performs graph reasoning with LLMs entirely in text space.

\subsection{Graph Foundation Models}

\noindent $\bullet$ \textbf{GFT}~\citep{DBLP:conf/nips/WangZCZ024}. This model treats computation trees as transferable patterns and builds a discrete tree vocabulary shared across graphs and tasks.

\subsection{Additional Baselines for Generative Link Prediction}

\noindent $\bullet$ \textbf{InstructGLM}~\citep{DBLP:conf/eacl/YeZWXZ24}. This framework verbalizes a node's multi-hop neighborhood into natural-language instructions and instruction-tunes generative LLMs to directly generate predictions for node classification and link prediction without GNNs.

\noindent $\bullet$ \textbf{KGT5-context}~\citep{kochsiek2023friendly}. This model extends the sequence-to-sequence knowledge graph completion model KGT5 by appending the query entity's one-hop neighborhood to the input sequence, and decodes the target entity directly as text.

\section{Implementation Details}
\label{app: implementation_details}

We conduct all experiments in PyTorch on four NVIDIA RTX A6000 GPUs. In the graph understanding module, we use OpenAI's GPT-4o mini to facilitate prompt-based GNNs. The number of GNN layers is selected from $\{1, 2, 3, 4\}$, and the first-order neighbor sampling ratio from $\{0.3, 0.6, 1.0\}$. In the graph inference module, we fine-tune either Flan-T5 or Llama3-8B with a learning rate of 3e-4 and a batch size of 4. To mitigate potential biases introduced by task-specific prompts, we design a prompt pool for each task and randomly sample prompts during instruction construction to enhance robustness. We adopt a standard early-stopping strategy during training: if the validation performance does not improve for a fixed number of consecutive epochs (determined per dataset), training is halted to prevent overfitting. All results are reported using 10-fold cross-validation, with mean performance and standard deviations. For evaluation, we use accuracy and Macro-F1 (M-F1) for node classification, and accuracy and HR@1 for link prediction. To ensure fair comparison, BERT is used for node initialization across all applicable baselines. For other hyperparameters of the compared methods, we follow the original papers and carefully tune them to each dataset.

\section{Prompt Design}
\label{Prompt Design}

This section provides templates of prompts, including prompt-based GNNs and task-oriented prompts.

\begin{tcolorbox}[width=\linewidth, colback=blue!5!white, colframe=blue!75!black, breakable, title={\textsc{Node Initialization Prompt}}]\label{prompt:node-initialization}

\small
The title of the paper is {\color{blue}<the title of the paper>}, the abstract of the paper is {\color{blue}<the abstract of the paper>}. Please summarize the paper.

\noindent{\textcolor{red}{/*** This prompt varies depending on the dataset. The present instance is for citation datasets only. ***/}}
\end{tcolorbox}

\begin{tcolorbox}[width=\linewidth, colback=blue!5!white, colframe=blue!75!black, breakable, title={\textsc{Layer-wise GNN Replication Prompt}}]\label{prompt:gnn-replication}

\small
Given the central node {\color{blue}<l-th round textual representation of the central node>}. The selected one-hop neighbors are [{\color{blue}<l-th round of node \#1>}, ..., {\color{blue}<l-th round node \#N>}].

Please aggregate neighbor nodes and update a concise yet meaningful representation for the central node. Note connected nodes should share similar semantics and vice versa.

\end{tcolorbox}

\begin{tcolorbox}[width=\linewidth, colback=blue!5!white, colframe=blue!75!black, breakable, title={\textsc{Node Classification Task Template}}]\label{prompt:node-classification}

\small

{Given \color{blue}<central node language-based ID>} has 1-hop connections with [..., {\color{blue}<1-hop neighbor language-based IDs>}, ...], and it also has 2-hop connections with [..., {\color{blue}<2-hop neighbor language-based IDs>}, ...].
Which category should {\color{blue}<central node language-based ID>} be classified as?

\end{tcolorbox}

\begin{tcolorbox}[width=\linewidth, colback=blue!5!white, colframe=blue!75!black, breakable, title={\textsc{Discriminative Link Prediction Task Template}}]\label{prompt:discriminative-link-prediction}

\small

{Given \color{blue}<central node language-based ID >} has 1-hop connections with [..., {\color{blue}<1-hop neighbor language-based IDs>}, ...], and it also has 2-hop connections with [..., {\color{blue}<2-hop neighbor language-based IDs>}, ...].
Among {\color{blue}<central node language-based ID>} and {\color{blue}<negative sampling node language-based ID>}, which node will be connected to {\color{blue}<central node language-based ID>}?

\end{tcolorbox}

\begin{tcolorbox}[width=\linewidth, colback=blue!5!white, colframe=blue!75!black, breakable, title={\textsc{Generative Link Prediction Task Template}}]\label{prompt:generative-link-prediction}

\small

{Given \color{blue}<central node language-based ID>} {\color{black} has 1-hop connections with [...,} {\color{blue}<1-hop neighbor language-based IDs>}{\color{black}, ...], and it also has 2-hop connections with [...,} {\color{blue}<2-hop neighbor language-based IDs>}{\color{black}, ...].}
Which node should be connected to {\color{blue}<central node language-based ID>}{\color{black}?}

\end{tcolorbox}

\section{Prompt-based GNN Analysis}
\label{Prompt-based GNN Analysis}

\begin{figure*}[!tb]
    \centering
    \includegraphics[width=0.92\textwidth]{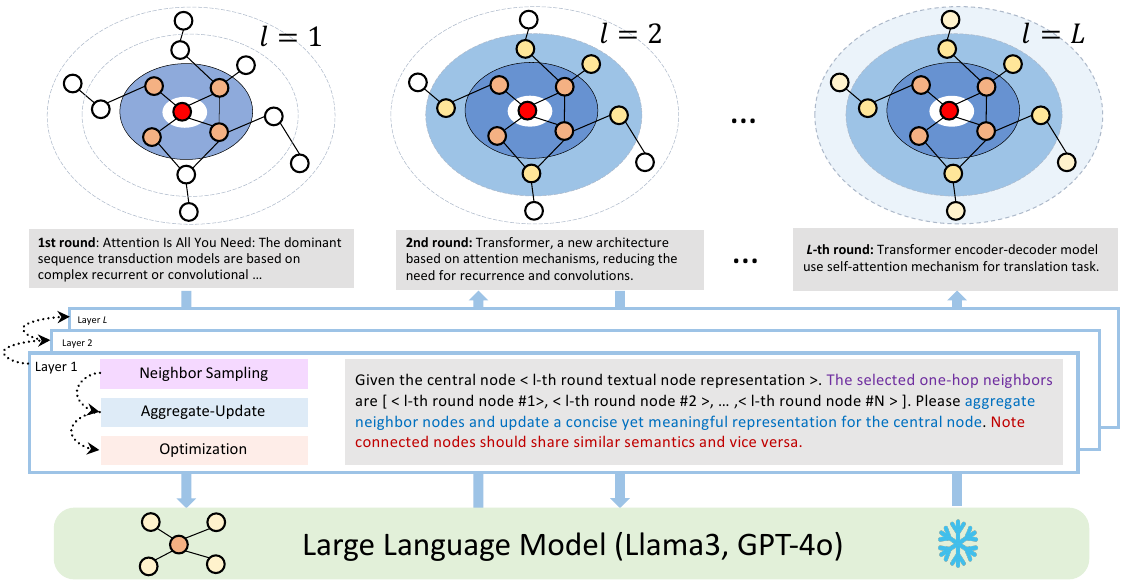}
    \caption{Graph understanding module via prompt-based GNNs. We prompt LLMs to achieve fine-grained reproduction of traditional GNN workflow, refining verbose textual representations into concise yet meaningful ones. In the prompt, neighbor sampling (see Equation~\ref{Neighbor Sampling.}) is in purple, the aggregation-update mechanism (see Equation~\ref{Agg-Update.}) in blue, and the optimization in red.}
    \label{Fig_GUM}
\end{figure*}

We illustrate the prompt-based GNN workflow in Figure~\ref{Fig_GUM} and validate it from two complementary angles: permutation sensitivity (\S\ref{sec: permutation}) and the layer-wise behavior of the prompt-based GNN (\S\ref{sec:repr_discrimination}), which covers structural discrimination, semantic retention, and the impact of GNN depth.

\subsection{Permutation Sensitivity}
\label{sec: permutation}
We evaluate the sensitivity of our prompt-based GNNs (using Flan-T5) to input order by comparing two variants: \textbf{(a) shuffle nodes}, which randomizes the order of neighbors, and \textbf{(b) shuffle tokens}, which injects position-specific tokens to nodes and shuffles them instead. As illustrated in Table~\ref{Permutation sensitivity of prompt-based GNNs.}, regardless of the shuffle method, both variants result in only minor fluctuations in node classification accuracy, indicating that prompt-based GNNs with the subsequent fine-tuning are robust to input permutation and maintain stable performance.

\begin{table}[!tb]
    \centering
    \caption{Permutation sensitivity of prompt-based GNNs.}
    \label{Permutation sensitivity of prompt-based GNNs.}
    \footnotesize
    \setlength{\tabcolsep}{12pt}
    \renewcommand{\arraystretch}{1.15}
    \begin{tabular}{c|c|c}
    \toprule
    Variant         & Cora             & Citeseer         \\
    \midrule
    shuffle nodes   & 90.64 $\pm$ 1.12 & 83.79 $\pm$ 0.89 \\
    shuffle tokens  & 90.55 $\pm$ 1.03 & 84.28 $\pm$ 1.06 \\
    \bottomrule
    \end{tabular}
\end{table}

\subsection{Layer-wise Behavior}
\label{sec:repr_discrimination}
\label{sec: self_text_retention}
\noindent \textbf{Structural Discrimination.}
To validate that our language-based graph vocabulary effectively preserves structural proximity, we conduct an analysis by encoding the final language-based IDs with a pretrained Sentence-BERT encoder and computing the cosine similarity between node pairs. The results on the Cora dataset are presented in Table~\ref{tab:cosine_similarity}. We observe that connected nodes exhibit a significantly higher mean similarity than disconnected nodes ($p < 10^{-5}$). Furthermore, the margin between the two groups grows steadily as the number of GNN layers increases, indicating stronger structural discrimination with deeper propagation. This trend highlights that our prompt-based GNN captures meaningful semantics with stacking aggregation-update layers, thereby suggesting that its representational behavior parallels that of traditional embedding-based GNNs. Notably, disconnected pairs remain around $0.03$ at every depth, indicating that deeper propagation sharpens structural discrimination without collapsing all nodes into a common representation.

\begin{table}[!tb]
    \centering
    \caption{Layer-wise mean cosine similarity of node embeddings on Cora, with margin reflecting the gap between connected and disconnected pairs.}
    \label{tab:cosine_similarity}
    \footnotesize
    \setlength{\tabcolsep}{8pt}
    \renewcommand{\arraystretch}{1.15}
    \begin{tabular}{l|c|c|c}
        \toprule
         & Connected & Disconnected & Margin \\
        \midrule
        Layer 1 & 0.171 & 0.036 & 0.135 \\
        Layer 2 & 0.223 & 0.032 & 0.191 \\
        Layer 3 & 0.308 & 0.034 & 0.274 \\
        \bottomrule
    \end{tabular}
\end{table}

\noindent \textbf{Semantic Retention and Integration.}
We next examine how each node balances its own semantics against the injected neighbor context. We encode the initial center-only summary and the output of each graph understanding layer with the same frozen encoder, and report the similarity to the initial summary (self-retention) and to the sampled neighbors before and after each layer-wise update. As shown in Table~\ref{tab: self_retention}, neighbor similarity consistently increases after every update, showing that structural context is progressively injected. Meanwhile, the representation stays closer to its own initial summary than to its neighbors at every depth, so central-node semantics are compressed and refined rather than erased. The two similarities nevertheless converge as layers are stacked, with the margin on Cora narrowing from $0.673$ at the first layer to $0.295$ at the fourth, indicating that the central node progressively loses its dominance.

\begin{table}[!tb]
    \centering
    \caption{Layer-wise self-retention and neighbor similarity of textual representations. Neighbor similarity is reported before and after each layer-wise update.}
    \label{tab: self_retention}
    \footnotesize
    \setlength{\tabcolsep}{3pt}
    \renewcommand{\arraystretch}{1.15}
    \begin{tabular}{l|cccc}
        \toprule
        Metric & $l=1$ & $l=2$ & $l=3$ & $l=4$ \\
        \midrule
        \multicolumn{5}{l}{\textit{Cora}} \\
        Self-retention & 0.837 & 0.758 & 0.687 & 0.632 \\
        Neighbor similarity (before) & 0.118 & 0.173 & 0.226 & 0.306 \\
        Neighbor similarity (after)  & 0.164 & 0.214 & 0.279 & 0.337 \\
        \midrule
        \multicolumn{5}{l}{\textit{Citeseer}} \\
        Self-retention & 0.807 & 0.724 & 0.643 & 0.584 \\
        Neighbor similarity (before) & 0.103 & 0.147 & 0.188 & 0.248 \\
        Neighbor similarity (after)  & 0.144 & 0.181 & 0.234 & 0.269 \\
        \bottomrule
    \end{tabular}
\end{table}

\begin{figure}[!tb]
    \centering
    \includegraphics[width=0.95\linewidth]{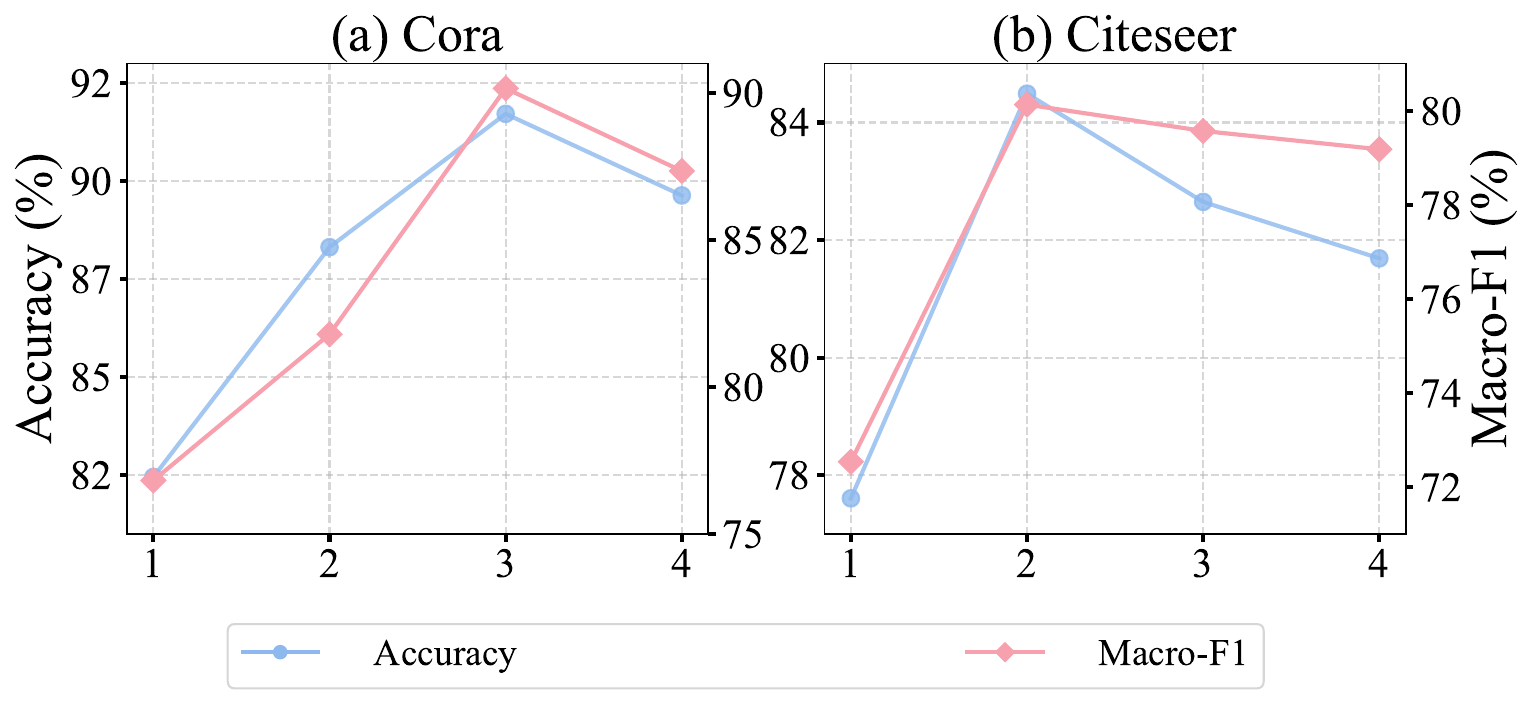}
    \caption{Impact of layer numbers.}
    \label{GNN_Layers}
\end{figure}

\label{sec: gnn_layers}
\noindent \textbf{Impact of GNN Depth.}
This trade-off is directly reflected in task performance.
As shown in Figure~\ref{GNN_Layers}, increasing the number of layers initially improves results by capturing broader context and higher-order dependencies. However, beyond a certain depth, performance plateaus or declines due to over-smoothing, a phenomenon also observed in traditional GNNs. The optimal depth is achieved with three layers on Cora and two layers on Citeseer. These results suggest that textual representations can be propagated across the graph in much the same way as numerical embeddings, effectively encoding both semantic and structural information.

\section{Graph Vocabulary Analysis}
\label{sec: vocabulary_analysis}

We evaluate the quality of our graph vocabulary using two metrics in Table~\ref{Vocabulary Analysis}.
\textbf{(a) Coverage.} We compute the proportion of nodes successfully assigned IDs. Since the graph vocabulary is grounded in the LLM's native vocabulary, any node's semantics can be expressed through a combination of natural language tokens. In practice, every node receives a valid ID, yielding full coverage across all datasets.
\textbf{(b) Uniqueness.} We measure the entropy of token distributions across nodes within both individual and merged graphs. Higher entropy indicates more diverse and fine-grained token usage, indicating stronger uniqueness and representational capacity. Notably, the merged vocabulary (``ALL'') achieves the highest entropy (12.2761), showing that integrating multiple graphs improves vocabulary expressiveness and discriminability, which reinforces the necessity of building GFMs.

\begin{table*}[!tb]
    \centering
    \caption{Quantitative graph vocabulary analysis.}
    \label{Vocabulary Analysis}
    \footnotesize
    \setlength{\tabcolsep}{16pt}
    \renewcommand{\arraystretch}{1.15}
    \begin{tabular}{c|cccc}
    \toprule
    Metric                 & Cora    & Citeseer & PubMed  & Ogbn-arxiv \\ \midrule
    \textit{Coverage (\%)} & 100.00  & 100.00   & 100.00  & 100.00     \\
    \textit{Entropy}       & 9.1116  & 9.4561   & 8.4643  & 10.7464    \\ \midrule[\heavyrulewidth]
    Metric                 & History & Photo    & WikiCS  & ALL        \\ \midrule
    \textit{Coverage (\%)} & 100.00  & 100.00   & 100.00  & 100.00     \\
    \textit{Entropy}       & 10.3752 & 11.3272  & 9.3825  & 12.2761    \\ \bottomrule
    \end{tabular}
\end{table*}

\section{Prompt Length and Scalability}
\label{Prompt Length Limitation Discussion}

In dense TAGs, high-degree nodes present a challenge for prompt-based GNNs, as their long prompts may exceed LLM input limits and hinder efficiency. To address this issue, we examine strategies that balance efficiency, scalability, and representational fidelity.

\begin{table*}[!tb]
\centering
\caption{Analysis of high-degree nodes and average connectivity.}
\label{tab: node_degree}
\footnotesize
\setlength{\tabcolsep}{8pt}
\renewcommand{\arraystretch}{1.15}
\begin{tabular}{l|ccccccc}
\toprule
Metric                    & Cora & Citeseer & PubMed & Ogbn-arxiv & History & Photo & WikiCS \\ \midrule
Avg. Degree               & 4.01 & 2.84     & 13.77  & 36.90      & 18.07   & 12.11 & 4.50   \\
High-degree ($>$300) Nodes & 0    & 0        & 0      & 0.20\%     & 0.25\%  & 0.19\% & 0     \\
\bottomrule
\end{tabular}
\end{table*}

\subsection{Neighbor Sampling in Scalable GNNs}
\label{sec: limitation_neighbor_sampling}
Following standard practices, we adopt a neighbor sampling strategy in each GNN layer to reduce prompt length and enhance robustness. As shown in Table~\ref{tab: sample_ratio}, sampling 60\% of neighbors yields the best overall performance across datasets. Meanwhile, limiting the maximum sampled neighbors is a common practice in scalable GNNs. For example, GraphSAGE samples 25 and 10 neighbors at the first and second hops, while PinSAGE~\citep{ying2018pinsage} selects from $\{10, 20, 50\}$. Following this rationale, we explore $\{10, 20, 30, 40\}$ and empirically set 20 as the cap, balancing LLM input constraints with sufficient structural coverage.

\subsection{Token Capacity of LLMs}
Our implementation leverages GPT-4o mini, which supports up to 128k input tokens per call. This capacity allows a single prompt to encode information from over 300 neighbors, given that each neighbor contributes about 400 tokens in the first propagation round. As shown in Table~\ref{tab: node_degree}, only a small fraction of nodes ($<$0.25\%) in a few datasets surpass this capacity. For these rare high-degree nodes, our neighbor sampling and truncation mechanism can bound prompt length, ensuring broad scalability to dense graphs.

\subsection{Prompt Design Considerations}
Prompt-based GNNs guide the LLM to aggregate and update neighbor information through structured prompts, producing increasingly compact yet expressive representations. Across layers, the average length shrinks from ${\sim}400$ to ${\sim}10$ tokens as semantic and structural information is progressively distilled. Since each node is processed independently and depends only on its local degree, prompt size remains bounded regardless of graph scale, ensuring both efficiency and scalability to large graphs.

\begin{table}[!tb]
\centering
\caption{The performance with varying neighbor sampling ratios across three datasets.}
\label{tab: sample_ratio}
\footnotesize
\setlength{\tabcolsep}{12pt}
\renewcommand{\arraystretch}{1.15}
\begin{tabular}{c|c|c|c}
\toprule
Sampling Ratio & 100\% & 60\% & 30\% \\
\midrule
Cora     & 81.36 & \color{purple}\textbf{89.35} & 88.36 \\
Citeseer & 79.34 & \color{purple}\textbf{82.23} & 81.13 \\
PubMed   & 90.23 & \color{purple}\textbf{91.12} & 89.34 \\
\bottomrule
\end{tabular}
\end{table}

\section{Adaptive Prompting for Cross-Domain Transfer}
\label{sec:adaptive_prompting}

While inter-domain transfer remains a community-wide challenge rather than a limitation of our method, the domain shift assumption inspires us to investigate adaptive prompting. We introduce a lightweight domain-adaptive strategy by prepending domain-specific instructions at inference time. For instance, during E-commerce$\rightarrow$Biomedical transfer, we add: \textit{``Now perform node classification/link prediction over a graph representing biomedical research citations, rather than co-purchase relations in e-commerce.''}
As summarized in Table~\ref{tab:adaptive-prompting}, adaptive prompting (\textit{w/ AP}) consistently improves results across all evaluation settings. These findings highlight adaptive prompting as an effective means of bridging domain gaps and a promising direction toward more generalizable GFMs.

\begin{table*}[!tb]
\centering
\caption{The effectiveness of adaptive prompting that alleviates inter-domain transfer challenges.}
\label{tab:adaptive-prompting}
\footnotesize
\setlength{\tabcolsep}{8pt}
\renewcommand{\arraystretch}{1.2}
\begin{tabular}{cc|cc|cc|cc|cc}
\toprule
\multicolumn{2}{c|}{\multirow{2}{*}{\textbf{Source}}} & \multicolumn{4}{c|}{\textbf{Pre-training}} & \multicolumn{4}{c}{\textbf{Co-training}} \\
\cmidrule(lr){3-6} \cmidrule(lr){7-10}
\multicolumn{2}{c|}{} & \multicolumn{2}{c}{Acc ($\uparrow$)} & \multicolumn{2}{c|}{M-F1 ($\uparrow$)} & \multicolumn{2}{c}{Acc ($\uparrow$)} & \multicolumn{2}{c}{M-F1 ($\uparrow$)} \\
\cmidrule(lr){3-4} \cmidrule(lr){5-6} \cmidrule(lr){7-8} \cmidrule(lr){9-10}
History & Photo & w/o & w/ & w/o & w/ & w/o & w/ & w/o & w/ \\
\midrule
\xmark & \xmark & 39.12 & 40.23 & 39.84 & 40.13 & 90.67 & 90.82 & 91.82 & 91.93 \\
\cmark & \xmark & 46.24 & 47.44 & 47.41 & 47.98 & 88.52 & 90.27 & 89.39 & 90.22 \\
\xmark & \cmark & 44.21 & 45.11 & 44.93 & 46.02 & 85.34 & 88.13 & 85.57 & 88.47 \\
\cmark & \cmark & 45.18 & 47.90 & 45.65 & 47.26 & 86.22 & 89.02 & 87.03 & 89.16 \\
\bottomrule
\end{tabular}
\end{table*}

\section{Dynamic Graph Adaptation}
\label{sec: dynamic_graphs}
PromptGFM adapts to evolving graphs through localized updates, without reprocessing the entire graph:

\noindent $\bullet$ \textbf{New node:} generate its initial textual representation and aggregate from its current neighbors.

\noindent $\bullet$ \textbf{New edge:} update the affected endpoint nodes and, optionally, their $L$-hop neighborhoods.

\noindent $\bullet$ \textbf{Changed node text:} regenerate only the changed node and the locally affected representations.

To assess this protocol, we conduct node classification on a dynamic DBLP co-authorship graph, comparing PromptGFM-dyn, a dynamic variant of PromptGFM equipped with the localized updates above, against representative GNN and GNN-LLM baselines under the same setting. As shown in Table~\ref{tab: dynamic_graphs}, PromptGFM-dyn consistently outperforms all baselines, suggesting that cached language-based IDs can be locally updated and remain effective in evolving graph settings.

\begin{table}[!tb]
    \centering
    \caption{Node classification on a dynamic DBLP co-authorship graph.}
    \label{tab: dynamic_graphs}
    \footnotesize
    \setlength{\tabcolsep}{10pt}
    \renewcommand{\arraystretch}{1.15}
    \begin{tabular}{l|c|c}
        \toprule
        Method & Accuracy & Macro-F1 \\
        \midrule
        GCN & 50.83 & 45.23 \\
        GAT & 50.28 & 45.39 \\
        GraphSAGE & 50.49 & 45.37 \\
        GraphPrompter & 52.37 & 47.10 \\
        OFA & 55.17 & 52.54 \\
        \midrule
        PromptGFM-dyn & \color{darkred}\textbf{59.46} & \color{darkred}\textbf{56.23} \\
        \bottomrule
    \end{tabular}
\end{table}

\section{Layer-wise Case Studies}
\label{Case Study}

We visualize the layer-by-layer refinement of node representations and identify which neighbor nodes contribute to each stage. Textual representations evolve from verbose descriptions into compact, high-density token sequences (8--10 tokens). After the first layer, nodes absorb neighbor semantics (e.g., ``error estimation''); after the second round, they integrate broader two-hop context (e.g., ``nonlinear regression analysis''). These observations demonstrate the explicit message-passing behavior of the prompt-based GNN in text space, effectively capturing both graph semantics and structural dependencies. We provide two representative cases below, from the Cora citation network and the PubMed biomedical citation network respectively.

\subsection{Case A: Cora}

\begin{tcolorbox}[width=\linewidth, colback=blue!5!white, colframe=blue!75!black, breakable, title={\textsc{Raw Text Attributes}}]

\small
Paper Title: Evaluating Neural Network Predictors by Bootstrapping. Abstract: We present a new method, inspired by the bootstrap, whose goal it is to determine the quality and reliability of a neural network predictor. Our method leads to more robust forecasting along with a large amount of statistical information on forecast performance that we exploit. We exhibit the method in the context of multi-variate time series prediction on financial data from the New York Stock Exchange. It turns out that the variation due to different resamplings (i.e., splits between training, cross-validation, and test sets) is significantly larger than the variation due to different network conditions (such as architecture and initial weights). Furthermore, this method allows us to forecast a probability distribution, as opposed to the traditional case of just a single value at each time step. We demonstrate this on a strictly held-out test set that includes the 1987 stock market crash. We also compare the performance of the class of neural networks to identically bootstrapped linear models.
\end{tcolorbox}

\begin{tcolorbox}[width=\linewidth, colback=blue!5!white, colframe=blue!75!black, breakable, title={\textsc{Round 0: Textual Feature Initialization}}]

\small
This method evaluates neural network predictors using a bootstrap-inspired approach, improving robustness in NYSE financial data forecasting. It predicts probability distributions rather than single values, surpassing resampling variations. Compared to bootstrapped linear models, it enhances accuracy, as demonstrated during the 1987 stock market crash.
\end{tcolorbox}

\begin{tcolorbox}[width=\linewidth, colback=blue!5!white, colframe=blue!75!black, breakable, title={\textsc{Round 1: 1st Layer Representations}}]

\small
This bootstrap method improves neural network evaluations in NYSE forecasting, {\color{red}comparable to traditional error estimation methods [A].} {\color{red}It clears and simplifies data structures [B]}, enhancing accuracy and predicting probability distribution.\\[2pt]
/*** {\color{red}[A]} comes from the paper: {\color{blue}\emph{A comparison of some error estimates for neural network models}}. ***/ \\[2pt]
/*** {\color{red}[B]} comes from the paper: {\color{blue}\emph{The Observer-Observation Dilemma in Neuro-Forecasting: Reliable Models From Unreliable Data Through Clearning}}. ***/
\end{tcolorbox}

\begin{tcolorbox}[width=\linewidth, colback=blue!5!white, colframe=blue!75!black, breakable, title={\textsc{Round 2: 2nd Layer Representations}}]

\small
This bootstrap-inspired method refines neural network evaluations for NYSE forecasting, highlighting {\color{red}local error bars from nonlinear regression analysis [C]}, improving accuracy with data simplification techniques.\\[2pt]
/*** {\color{red}[C]} comes from the paper: {\color{blue}\emph{Learning Local Error Bars for Nonlinear Regression}}. ***/
\end{tcolorbox}

\begin{tcolorbox}[width=\linewidth, colback=blue!5!white, colframe=blue!75!black, breakable, title={\textsc{Round 3: 3rd Layer Representations}}]

\small
Bootstrap method refines neural forecasts, integrates error bars, enhances accuracy.
\end{tcolorbox}

\subsection{Case B: PubMed}

\begin{tcolorbox}[width=\linewidth, colback=blue!5!white, colframe=blue!75!black, breakable, title={\textsc{Raw Text Attributes}}]

\small
Paper Title: Insulin dependent diabetes in Asians. Abstract: Type 1 diabetes is said to be extremely rare in children in India, where diabetes treated with insulin may be due to chronic pancreatic disease or malnutrition. To see whether typical type 1 diabetes occurred in Asian children in the United Kingdom, all known Asian children with diabetes in industrial West Yorkshire were ascertained. A total of 17 such children were studied; of these, seven were from three multiplex families and two fathers from these families had diabetes. All children were ketosis prone and developed diabetes while resident in the UK. There were significant increases in HLA-B8 and HLA-DR3 and increases in HLA-DR4 and HLA-DR3/DR4, while HLA-B15 was absent. Islet cell antibodies, either IgG or complement fixing, were present in four of 18 subjects tested, all of whom had disease of short duration. The prevalence of type 1 diabetes in Asian children aged 15 years or less in West Yorkshire was 36/100,000, assuming complete ascertainment. It is concluded that typical type 1 diabetes may occur in Asian children and this condition may be more common in families who have migrated to the UK.
\end{tcolorbox}

\begin{tcolorbox}[width=\linewidth, colback=blue!5!white, colframe=blue!75!black, breakable, title={\textsc{Round 0: Textual Feature Initialization}}]

\small
The study examined type 1 diabetes prevalence in 17 Asian children in West Yorkshire, UK, focusing on family history. Seven were from multiplex families, and all were ketosis-prone. Significant increases in HLA-B8, HLA-DR3, and HLA-DR4 were found, with no HLA-B15. Islet cell antibodies were present in four of 18 subjects. The diabetes prevalence was 36/100,000, suggesting it may be more common in migrant families.
\end{tcolorbox}

\begin{tcolorbox}[width=\linewidth, colback=blue!5!white, colframe=blue!75!black, breakable, title={\textsc{Round 1: 1st Layer Representations}}]

\small
The study examined type 1 diabetes in Asian children in West Yorkshire, UK. {\color{red}Environmental factors and migration may raise incidence [A]}, unlike the {\color{red}low incidence of insulin-dependent diabetes in Karachi [B]}. Increases in HLA-B8, HLA-DR3, and HLA-DR4 were found.\\[2pt]
/*** {\color{red}[A]} comes from the paper: {\color{blue}\emph{Evidence for an environmental effect in the aetiology of insulin dependent diabetes in a transmigratory population}}. ***/ \\[2pt]
/*** {\color{red}[B]} comes from the paper: {\color{blue}\emph{Incidence of insulin dependent diabetes mellitus in Karachi, Pakistan}}. ***/
\end{tcolorbox}

\begin{tcolorbox}[width=\linewidth, colback=blue!5!white, colframe=blue!75!black, breakable, title={\textsc{Round 2: 2nd Layer Representations}}]

\small
The study examined type 1 diabetes in Asian children in West Yorkshire, UK. {\color{red}Migration may raise incidence [C]}, unlike the low incidence in Karachi. Increases in HLA markers were found, similar to {\color{red}North and South Indian diabetics [D]}.\\[2pt]
/*** {\color{red}[C]} comes from the paper: {\color{blue}\emph{HLA-DR antigen frequencies in a North Indian type I diabetic population}}. ***/ \\[2pt]
/*** {\color{red}[D]} comes from two papers: {\color{blue}\emph{HLA-DR antigen frequencies in a North Indian type I diabetic population}} and {\color{blue}\emph{HLA, complement C2, C4, properdin factor B and glyoxalase types in South Indian diabetics}}. ***/
\end{tcolorbox}

\begin{tcolorbox}[width=\linewidth, colback=blue!5!white, colframe=blue!75!black, breakable, title={\textsc{Round 3: 3rd Layer Representations}}]

\small
Type 1 diabetes. Migration and HLA markers linked to increased diabetes incidence.
\end{tcolorbox}

\end{document}